\documentclass[11pt]{article}

\usepackage[preprint]{acl}
\usepackage{times}
\usepackage{latexsym}
\usepackage[T1]{fontenc}
\usepackage[utf8]{inputenc}
\usepackage{microtype}
\usepackage{inconsolata}
\usepackage{graphicx}
\usepackage{amsmath}
\usepackage{amssymb}
\usepackage{subcaption}
\usepackage[skins]{tcolorbox}
\usepackage{pifont}
\usepackage{multirow}
\usepackage{booktabs}
\usepackage{tabularx}
\usepackage{array}
\usepackage{amssymb}
\usepackage{makecell}
\usepackage{xspace}
\definecolor{retrievalbg}{HTML}{FDF6E3}
\definecolor{modelbg}{HTML}{E8F4F8}
\definecolor{successbg}{HTML}{EBF3E6}
\definecolor{errorbg}{HTML}{F9EBEA}
\tcbset{
    mybox/.style={
        colframe=gray!50, boxrule=0.5pt, arc=2pt, 
        left=4pt, right=4pt, top=4pt, bottom=4pt,
        fonttitle=\bfseries\small, fontupper=\small
    }
}

\newcommand{\up}[1]{\,\textcolor{blue}{{\scriptsize(+#1)}}}
\newcommand{\down}[1]{\,\textcolor{red}{{\scriptsize(-#1)}}}
\newcommand{\same}{\,{\scriptsize(0.000)}}
\newcommand{\emojiname}{\raisebox{-0.2\baselineskip}{\includegraphics[height=1.1\baselineskip]{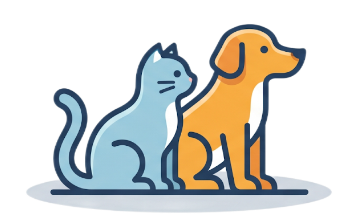}}\textsc{PetQA}}

\title{\emojiname: Benchmarking Veterinary Knowledge and Clinical Reasoning}

\author{
\textbf{Taegyun Kim}\textsuperscript{\(\spadesuit\)} \quad
\textbf{Youngwook Ham}\textsuperscript{\(\clubsuit\)} \quad
\textbf{Jungwook Rhim}\textsuperscript{\(\clubsuit\)} \\
\textbf{Ju-Hyun An}\textsuperscript{\(\clubsuit\)} \quad
\textbf{Sungkyu Park}\textsuperscript{\(\heartsuit\)} \quad
\textbf{Kunwoo Park}\textsuperscript{\(\diamondsuit\spadesuit\)} \\
\textsuperscript{\(\spadesuit\)}Department of Intelligent Semiconductors, Soongsil University\\
\textsuperscript{\(\clubsuit\)}Kangwon National University \\
\textsuperscript{\(\heartsuit\)}KDI School of Public Policy and Management\\
\textsuperscript{\(\diamondsuit\)}School of AI Convergence, Soongsil University\\
\texttt{\small tgkim12@soongsil.ac.kr, shaun@kdischool.ac.kr, kunwoo.park@ssu.ac.kr}
}

\newcommand{\mydata}[0]{\textsc{PetQA}\xspace}
\newcommand{\mydatabench}[0]{\textsc{PetQA-Bench}\xspace}

\begin{document}
\maketitle

\begin{abstract}
We introduce \mydata, a Korean long-form question-answering (QA) benchmark for evaluating veterinary knowledge and clinical reasoning in large language models (LLMs) and large vision-language models (LVLMs). \mydata contains 10,076 text-only and 8,751 multimodal QA pairs derived from real-world questions about dogs and cats, paired with answers from expert veterinarians. Its test split, \mydatabench, further includes annotations for question types and clinical conditions. We evaluate eighteen models using ROUGE, BERTScore, and LLM-as-a-judge metrics for factuality and helpfulness under three settings: zero-shot inference, retrieval-augmented generation (RAG), and supervised fine-tuning (SFT). The benchmarking results provide an overview of the strengths and limitations of current models in addressing veterinary clinical queries and highlight the need for more effective adaptation methods to develop clinically reliable AI systems for veterinary care. To facilitate broader use, we additionally provide translated versions of \mydatabench in five languages.
\end{abstract}

\begin{table*}[t]
\centering
\small
\resizebox{\textwidth}{!}{
\begin{tabular}{lcccccc}
\toprule
\textbf{Dataset} & \textbf{Source} & \textbf{Answer Format} & \textbf{Language} & \textbf{Modality} & \textbf{Size} & \textbf{Target Subject} \\ \midrule
BioASQ~\cite{tsatsaronis2015overview} & \multirow{2}{*}{\makecell{Medical \\ Article}} & \multirow{2}{*}{\makecell{Span-based,\\Binary}} & En & Text & 4,721 & \multirow{12}{*}{\shortstack{Human}} \\
PubMedQA~\cite{jin2019pubmedqa} &  &  & En & Text & 1,000 &  \\ \cmidrule(r){1-6}
MMLU (Med.)~\cite{hendrycks2021measuring} & \multirow{8}{*}{\makecell{Medical\\Licensing\\Examination}} & \multirow{8}{*}{\shortstack{Multiple-choice}} & En & Text & 1,089 &  \\
MedQA~\cite{jin2021disease} &  &  & En, Zh & Text & 61,097 &  \\
MedMCQA~\cite{pal2022medmcqa} &  &  & En & Text & 193,155 &  \\
CMExam~\cite{liu2023benchmarking} &  &  & Zh & Text & 68,119 &  \\
MedBench~\cite{liu2024medbench} &  &  & Zh & Text & 40,041 &  \\
KorMedMCQA~\cite{kweon2024kormedmcqa} &  &  & Ko & Text & 7,469 &  \\
MLEC-QA~\cite{li2021mlec} &  &  & Zh & Multimodal & 136,236 &  \\
MedXpertQA~\cite{zuo2025medxpertqa} &  &  & En & Multimodal & 4,460 &  \\ \cmidrule(r){1-6}
K-QA~\cite{manes2024k} & \multirow{2}{*}{\shortstack{Online Platform}} & \multirow{2}{*}{\makecell{Open-ended\\(long-form)}} & En & Text & 1,212 &  \\
\citet{hosseini2024a} &  &  & En & Text & 1,077 &  \\
\midrule
\multirow{2}{*}{\textbf{\mydata (Ours)}} & \multirow{2}{*}{Online Platform} & \multirow{2}{*}{\makecell{Open-ended\\(long-form)}} & \multirow{2}{*}{Ko} & \multirow{2}{*}{Multimodal} & \multirow{2}{*}{18,827} & \multirow{2}{*}{\makecell{Companion Animals\\(Dogs and Cats)}} \\
 &  &  &  &  &  &  \\ \bottomrule
\end{tabular}
}
\caption{Comparison of \mydata with existing medical benchmarks.}
\label{tab:medical_benchmark_dataset_comparison}
\end{table*}

\section{Introduction}
With advances in LLMs and LVLMs, their adoption in high-stakes domains such as healthcare has grown substantially~\cite{jeong2024medical, singhal2023large, li2023llavamed}. Despite the increasing number of households with companion animals and the growing demand for pet care and health management~\cite{appa2025state}, most existing benchmarks for medical AI evaluation have primarily focused on human medicine~\cite{jin2022biomedical, wang2025trustworthy}, leaving limited resources for evaluating clinical knowledge and reasoning capabilities in specialized domains such as veterinary medicine~\cite{luo2025empec}. Pet owners often seek advice from AI systems about abnormal symptoms observed in their companion animals~\cite{rspca2026kindness}, highlighting the need to systematically assess models' veterinary knowledge and clinical reasoning capabilities in realistic scenarios. 

To address this gap, we present \mydata, a Korean long-form QA dataset for veterinary medicine, comprising 10,076 QA pairs with text-only questions and 8,751 QA pairs with multimodal questions. We collect diverse real-world questions about dogs and cats from a major QA platform in South Korea, along with high-quality answers provided by verified experts. The data undergo a series of preprocessing steps and a quality assessment conducted with a veterinary expert holding a Ph.D. The test split, referred to as \mydatabench, contains 2,000 QA pairs per modality and includes annotations for question types and clinical conditions. The combination of a large-scale collection of real-world questions, long-form reference answers from veterinary experts, and both text-only and multimodal questions enables the systematic evaluation of LLMs' and LVLM's veterinary knowledge and clinical reasoning abilities in realistic scenarios. \looseness=-1

For initial benchmarking, we evaluate eighteen models---categorized as closed LVLMs, open-weight LVLMs, and open-weight LLMs---using two traditional metrics (ROUGE and BERTScore) and two LLM-as-a-judge metrics (factuality and helpfulness). ROUGE, BERTScore, and factuality are reference-based metrics evaluated against veterinary expert-provided answers, whereas helpfulness is assessed in a reference-free manner. We examine zero-shot inference, retrieval-augmented generation (RAG), and supervised fine-tuning (SFT) as three model evaluation settings. We find that closed models generally outperform open-weight models, particularly in factuality and helpfulness; all models perform consistently worse on multimodal questions than on text-only questions; and RAG and SFT yield inconsistent improvements in LLM-as-a-judge metrics. These findings characterize the capabilities and limitations of current models for veterinary clinical QA and underscore the need for more effective methods to enable more reliable AI support for veterinary care.

Our contributions are summarized as follows:
\begin{itemize}
    \item We introduce \mydata, a Korean long-form QA dataset designed to evaluate the ability of LLMs and LVLMs to respond to veterinary clinical queries. To the best of our knowledge, \mydata is the first long-form QA dataset for veterinary medicine.
    \item We present initial benchmarking results for eighteen models using two traditional and two LLM-as-a-judge metrics under zero-shot, RAG, and SFT settings. 
    \item We publicly release \mydata through our GitHub repository\footnote{\url{https://github.com/ssu-humane/PetQA}}. To facilitate broader research in the NLP community, we additionally provide translated versions of \mydatabench in five languages.
\end{itemize}

\begin{figure*}[t]
    \centering
    \includegraphics[width=\textwidth]{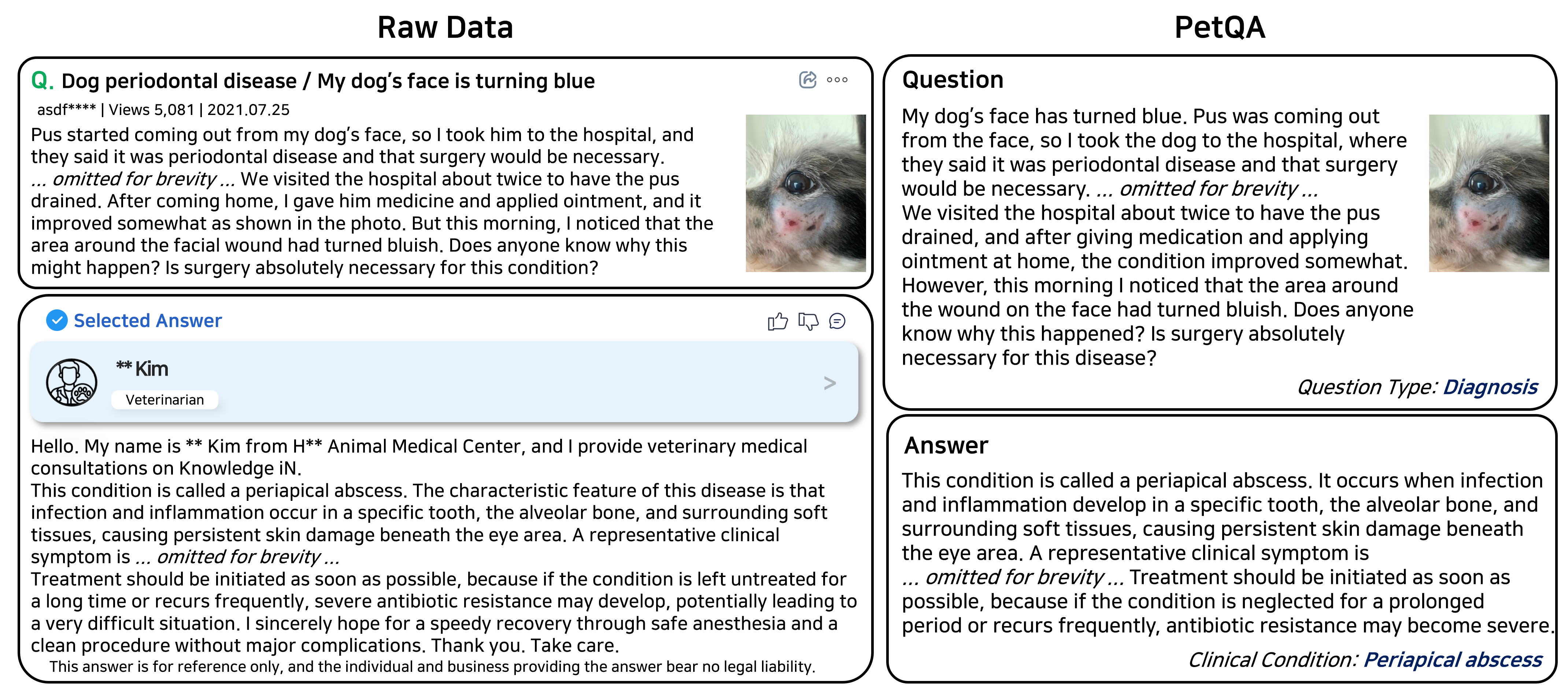}
    \caption{A raw data sample (left) and its corresponding example from \mydata (right).}
    \label{fig:dataset_process}
\end{figure*}

\section{Related Work}
Medical question-answering benchmarks can be categorized by question source and answer format. PubMedQA~\cite{jin2019pubmedqa} and BioASQ~\cite{tsatsaronis2015overview} use medical article-based questions with span-based or binary answers, whereas MMLU (Med.)~\cite{hendrycks2021measuring}, MedQA~\cite{jin2021disease}, and MedMCQA~\cite{pal2022medmcqa} use multiple-choice questions based on medical licensing examinations.

Multiple-choice question answering (MCQA) benchmarks~\cite{kweon2024kormedmcqa, zuo2025medxpertqa} are widely used because they enable straightforward quantitative evaluation of medical knowledge. 
However, recent studies suggest that MCQA may overestimate clinical reasoning ability, as models can exploit superficial patterns among answer choices rather than demonstrate genuine understanding~\cite{griot2025pattern}. Moreover, physicians are not provided with predefined options when making clinical decisions~\cite{cocchieri2026remedqa}. 
To address these limitations, recent studies have explored long-form QA, which requires detailed answers to open-ended questions~\cite{fan2019eli5}. \citet{manes2024k} constructed K-QA, comprising 1,212 patient questions and 201 curated physician answers, and introduced NLI-based metrics for clinical comprehensiveness and hallucination. Similarly, \citet{hosseini2024a} introduced a benchmark with 1,077 real-world consumer queries and long-form answers evaluated by medical doctors. 

Existing medical QA benchmarks have largely focused on human healthcare and have been primarily developed in English and Chinese~\cite{li2021mlec,liu2023benchmarking,liu2024medbench}. 
In veterinary medicine, benchmark resources remain scarce, with existing benchmarks limited to MCQA~\cite{luo2025empec,tam-etal-2026-vistw}.
To bridge this gap, we present \mydata, a Korean long-form QA benchmark for veterinary medicine that enables comprehensive evaluation of QA systems for dogs and cats, two of the most common companion animals in veterinary practice. A comparison between \mydata and existing medical QA resources is provided in Table~\ref{tab:medical_benchmark_dataset_comparison}.

\section{Dataset: \mydata}
We introduce \mydata, a long-form QA dataset designed to benchmark the veterinary knowledge and clinical reasoning capabilities of LLMs and LVLMs. \mydata comprises both text-only and multimodal questions about dogs and cats in realistic clinical scenarios, with answers provided by expert veterinarians. Figure~\ref{fig:dataset_process} illustrates how a raw data sample was processed and labeled.
Further dataset details are provided in Appendix~\ref{appendix:dataset_details}.

\subsection{Data Collection and Preprocessing}
\label{sec:data_collection_preprocessing}

\paragraph{Data Collection}
We collected question-answer pairs related to veterinary care from Naver Knowledge iN, one of the major community-driven online QA platforms in South Korea. Users can post questions to topic-specific boards and receive answers. Among multiple answers, the questioner can select one or more as helpful answers. The platform also provides expert answers through collaborations with affiliated organizations; users who possess nationally authorized professional licenses and association memberships can participate as experts~\cite{choi2025chatgpt}. Verification status is displayed through badges.

The target board for pet medical consultation contains a wide range of pet health inquiries, from general veterinary knowledge to real-world clinical questions such as symptom diagnosis and disease treatment. Each post consists of a question title, a question body that may include images, and a set of answers. Our initial data collection comprised 83,509 posts published between 2014 and 2024, each containing at least one answer selected as helpful by the questioner. The answers were provided either by anonymized users without verification or by 23 experts, all of whom are veterinarians whose identities were verified by the platform.

\paragraph{Rule-based Preprocessing}
We focused on dogs and cats because they are the two most common companion animals both globally~\cite{hoffmann2018empirical} and on the platform. After filtering out posts with duplicate questions or corrupted images, we retained 27,124 posts about the target animals, with answers selected by verified experts. Based on our preliminary investigation and preprocessing practices adopted in previous QA research~\cite{nguyen2023medredqa,wang2026refact}, we applied additional filtering and preprocessing steps to these posts.

\paragraph{LLM-based Preprocessing}
Following recent work~\cite{arias2025automatic, baumgartner2025peerqa}, we leveraged an LLM-based preprocessing pipeline to construct a coherent set of QA pairs while reducing noisy expressions. We first prompted GPT-4o-mini~\cite{hurst2024gpt} to filter out irrelevant posts, including questions unrelated to pet medical consultation (e.g., seeking advice on admission to veterinary school), answers based on unsupported speculation, and uninformative answers such as generic recommendations to visit a clinic. We further sanitized the text by removing personally identifiable information, correcting grammatical and spelling errors, and removing promotional content.

To assess the reliability of the LLM-based preprocessing, we manually evaluated 100 randomly sampled instances using two criteria: \emph{coherence} and \emph{completeness}. The results showed that 97 samples were coherent and 92 preserved all essential information without omission. The prompts and detailed evaluation guidelines are provided in Appendix~\ref{appendix:prompts} and \ref{appendix:llm-based_preprocessing}, respectively.

\paragraph{Data Split}
We split both the text-only and multimodal subsets into training, validation, and test sets, following the configurations of existing resources (Table~\ref{tab:medical_benchmark_dataset_comparison}). 
\begin{itemize}
    \item \textsf{Text}: Text-only questions with text answers (6,076 / 2,000 / 2,000)
    \item \textsf{Multimodal}: Questions with an image and text answers (4,751 / 2,000 / 2,000)
\end{itemize}
The test sets, collectively referred to as \mydatabench, are used primarily for benchmarking experiments, whereas the training and validation sets are used for supervised fine-tuning experiments. 

\subsection{Label Annotation}
\label{sec:annotation}
For \mydatabench, we annotated each questions with two labels: question type and clinical condition. These labels were designed to assess model performance across different clinical scenarios and to determine whether model responses contain key clinical conditions, respectively. 

The first step involved constructing a reliable annotation scheme and training the annotators. The initial guideline was carefully reviewed by the fourth author, a veterinary expert holding a Ph.D. Using 50 randomly sampled examples from the test set, six annotators participated in a pilot task in which they labeled the samples according to the initial guideline. Inter-annotator agreement was measured using Krippendorff's $\alpha$~\cite{krippendorff2018content}, and the process was repeated until $\alpha$ exceeded 0.7. This indicated that the refined guideline yielded reliable annotations. When agreement was low, the annotators and authors discussed disagreements to resolve them and refine the guideline. 

In the second step, the remaining samples in the test set were divided among the six annotators and independently labeled. We hired the annotators from the authors' institution and paid them in accordance with local wage laws. All annotators were students majoring in AI (four master's students and two undergraduate students), and two had experience caring for dogs and/or cats.

\paragraph{Question Type}
Each question was assigned a single label based on the perceived intent of the questioner. Following prior research in medical QA~\cite{zuo2025medxpertqa}, we established four categories:
(1) \textit{Diagnosis}, which includes the identification of diseases based on symptoms and the inference of possible causes;
(2) \textit{Treatment}, which includes treatment methods and preventive measures;
(3) \textit{Basic veterinary knowledge}, which covers general knowledge such as medical concepts and disease mechanisms;
(4) \textit{Miscellaneous}, which includes questions unrelated to pet medical consultation or administrative queries (e.g., hospital information and costs).

\paragraph{Clinical Condition}

For questions labeled as \emph{diagnosis}, annotators labeled the corresponding clinical conditions based on the selected answer. Clinical conditions include diseases, syndromes, and clinical states~\cite{tresker2020typology}. To support the annotation process, we provided a deduplicated list of 5,378 clinical-condition entries compiled from three reputable sources, including an official animal disease classification system~\cite{mafra-animal-disease}. Entries include conditions such as \emph{vomiting} and \emph{diarrhea}.

\begin{table}[t]
\centering
\resizebox{0.9\linewidth}{!}{%
\begin{tabular}{lrr}
\toprule
\textbf{Metric} & \textsf{Text} & \textsf{Multimodal}\\
\midrule
\textbf{Target species} & & \\
Dog & 1,672 & 1,650 \\
Cat & 328 & 350 \\
\midrule
\multicolumn{3}{l}{\textbf{Question type}} \\
Diagnosis & 1,231 & 1,699 \\
Treatment & 377 & 129 \\
Basic veterinary knowledge & 221 & 60 \\
Miscellaneous & 171 & 112 \\
\midrule
\textbf{Text tokens} & & \\
\emph{Questions}\\
- Max. & 838.0 & 819.0 \\
- Mean. & 143.6 & 104.3 \\
- Min. & 10.0 & 9.0 \\
\emph{Answers} \\
- Max. & 430.0 & 796.0 \\
- Mean. & 119.6 & 141.7 \\
- Min. & 11.0 & 16.0 \\
\bottomrule
\end{tabular}}
\caption{Descriptive statistics of \mydatabench.}
\label{tab:petqa_statistics}
\end{table}

\begin{figure}[t]
    \centering
    \includegraphics[width=\columnwidth]{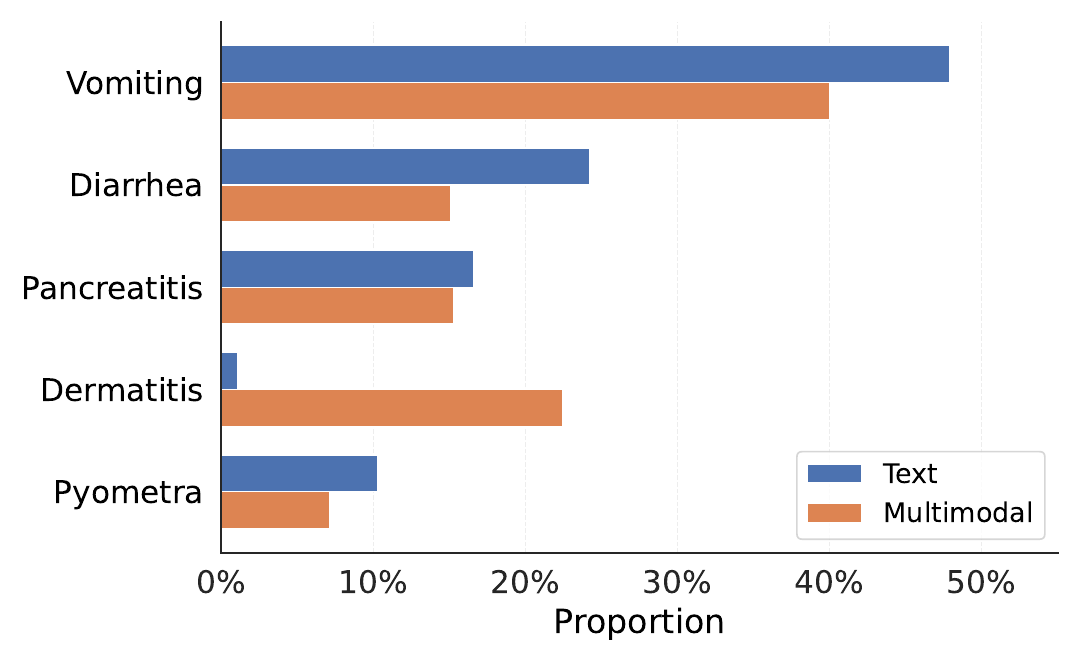}
    \caption{Distribution of the top-five clinical conditions in \mydatabench.}
    \label{fig:clinical_condition_distribution}
\end{figure}

\paragraph{Quality Check}
The fourth author, a veterinary expert, reviewed 100 randomly sampled QA pairs and their labels, with 50 from each of the \textsf{Text} and \textsf{Multimodal} sets. The evaluation used two criteria: \emph{relevance}, which assesses whether the answer appropriately addresses the question, and \emph{correctness}, which assesses whether the answer is factually accurate. The results showed that 99 samples were relevant and 85 were correct, supporting the dataset's overall reliability. Further details of the expert verification are provided in Appendix~\ref{appendix:expert_verification}.

\subsection{Descriptive Analysis}
\label{subsec:descriptive_analysis}
Table~\ref{tab:petqa_statistics} shows the descriptive statistics of \mydatabench. Questions about dogs are more frequent than those about cats. In terms of question type, \emph{Diagnosis} is the most frequent category, followed by \emph{Treatment}, indicating that our benchmark contains a substantial proportion of queries involving clinical scenarios. In particular, \emph{Diagnosis} is the most frequent category in both the \textsf{Text} and \textsf{Multimodal} sets, with a higher proportion in the \textsf{Multimodal} set. This trend suggests that images may help questioners convey symptoms.

Figure~\ref{fig:clinical_condition_distribution} shows the distributions of the five most frequent clinical conditions in \mydatabench, with proportions calculated separately for the \textsf{Text} and \textsf{Multimodal} sets. Vomiting is the most prevalent condition in both sets, followed by diarrhea and pancreatitis. In contrast, dermatitis ranks among the most frequent clinical conditions primarily in the \textsf{Multimodal} set. The \textsf{Text} and \textsf{Multimodal} sets cover 243 and 233 unique conditions, respectively, and 352 unique conditions when combined. These results highlight the broad range of clinical conditions represented in our dataset.

\subsection{Multilingual Extensions}
To improve the dataset's global applicability and broaden its potential impact, we translated \mydatabench into five widely spoken languages---English, German, Chinese, Indonesian, and Arabic---using an LLM-based machine translation pipeline. We selected these target languages based on their broad speaker coverage and linguistic diversity~\cite{schneider-sitaram-2024-m5}. We used GPT-4o-mini with a translation prompt adapted from \citet{lee2025koblex}.
Further details on the translation process and analysis are provided in Appendix~\ref{appendix:additional_results}.

\section{Experiments}

This section describes the experimental configuration, target models, and evaluation metrics used in the benchmarking experiments.

\subsection{Configuration}
Using \mydatabench, we assessed the veterinary knowledge and clinical reasoning capabilities of LLMs and LVLMs under three settings. For closed-book QA, we evaluated their \emph{zero-shot} abilities based on the knowledge encoded in their parameters. For open-book QA, we employed a RAG~\cite{lewis2020retrieval} pipeline using a veterinary reference widely used in clinical practice as the knowledge source. For SFT, we trained the models using the training set of \mydata. We used greedy decoding with a temperature of 0 for all models, resulting in deterministic outputs for open-weight models. For closed models, we reported scores averaged over three runs. Further details on the experimental configurations are provided in Appendix~\ref{appendix:experimental_setups}.

\subsection{Models}
We evaluated eighteen models that support Korean. Based on their vision capabilities and weight availability, we categorized the models into three groups: (1) \textbf{Closed LVLMs}: GPT-4.1 mini, GPT-4.1, Gemini 2.5 Flash, and Gemini 2.5 Pro; (2) \textbf{Open-weight LVLMs}: Qwen3-VL-8B, Qwen3-VL-32B, Gemma-3-12B, Gemma-3-27B, MedGemma-27B, A.X-4.0-VL-Light, and HCX-SEED-Vision-3B; and (3) \textbf{Open-weight LLMs}: Qwen3-8B, Qwen3-32B, Gemma-2-9B, Gemma-2-27B, MedGemma-27B-Text, A.X-4.0-Light, and HCX-SEED-Text-1.5B. Model checkpoints and additional details are provided in Appendix~\ref{appendix:experimental_setups}.

\begin{table*}[t]
\centering
\resizebox{\textwidth}{!}{
    \begin{tabular}{lcccccccc}
    \toprule
    \multirow{2}{*}{Model} & \multicolumn{4}{c}{\textsf{Text}} & \multicolumn{4}{c}{\textsf{Multimodal}} \\
    \cmidrule(lr){2-5} \cmidrule(lr){6-9}
    & ROUGE & BERTScore & Factuality & Helpfulness & ROUGE & BERTScore & Factuality & Helpfulness \\
    \midrule
    
    \multicolumn{9}{c}{\textit{Closed large vision-language models}} \\
    \midrule
    GPT-4.1 mini 
    & \textbf{0.303}{\scriptsize\,($\pm$ 0.001)} & \textbf{0.746}{\scriptsize\,($\pm$ 0.000)} & \underline{0.551}{\scriptsize\,($\pm$ 0.002)} & 4.728{\scriptsize\,($\pm$ 0.002)}
    & 0.237{\scriptsize\,($\pm$ 0.001)} & \textbf{0.726}{\scriptsize\,($\pm$ 0.000)} & \textbf{0.531}{\scriptsize\,($\pm$ 0.002)} & 4.597{\scriptsize\,($\pm$ 0.005)} \\
    GPT-4.1 
    & 0.290{\scriptsize\,($\pm$ 0.000)} & \underline{0.743}{\scriptsize\,($\pm$ 0.000)} & \textbf{0.569}{\scriptsize\,($\pm$ 0.001)} & 4.880{\scriptsize\,($\pm$ 0.003)} 
    & 0.234{\scriptsize\,($\pm$ 0.001)} & \textbf{0.726}{\scriptsize\,($\pm$ 0.001)} & \textbf{0.531}{\scriptsize\,($\pm$ 0.002)} & 4.715{\scriptsize\,($\pm$ 0.042)} \\
    Gemini 2.5 Flash 
    & 0.279{\scriptsize\,($\pm$ 0.000)} & 0.731{\scriptsize\,($\pm$ 0.000)} & 0.541{\scriptsize\,($\pm$ 0.001)} & \underline{4.914}{\scriptsize\,($\pm$ 0.007)} 
    & 0.224{\scriptsize\,($\pm$ 0.001)} & 0.709{\scriptsize\,($\pm$ 0.000)} & 0.495{\scriptsize\,($\pm$ 0.004)} & 4.679{\scriptsize\,($\pm$ 0.011)} \\
    Gemini 2.5 Pro 
    & 0.280{\scriptsize\,($\pm$ 0.001)} & 0.738{\scriptsize\,($\pm$ 0.000)} & 0.550{\scriptsize\,($\pm$ 0.002)} & \textbf{4.923}{\scriptsize\,($\pm$ 0.000)} 
    & \textbf{0.246}{\scriptsize\,($\pm$ 0.000)} & \textbf{0.726}{\scriptsize\,($\pm$ 0.001)} & \underline{0.524}{\scriptsize\,($\pm$ 0.008)} & \textbf{4.814}{\scriptsize\,($\pm$ 0.012)} \\
    \midrule
    \multicolumn{9}{c}{\textit{Open-weight large vision-language models}} \\
    \midrule
    Qwen3-VL-8B & 0.271 & 0.737 & 0.499 & 3.961 & 0.204 & 0.721 & 0.461 & 3.076 \\
    Qwen3-VL-32B & 0.258 & 0.730 & 0.513 & 4.867 & 0.215 & 0.712 & 0.489 & \underline{4.794} \\
    Gemma-3-12B & 0.287 & 0.740 & 0.510 & 4.561 & 0.243 & 0.723 & 0.490 & 4.255 \\
    Gemma-3-27B & 0.280 & 0.737 & 0.519 & 4.870 & \underline{0.244} & \underline{0.724} & 0.496 & 4.626 \\
    MedGemma-27B & 0.259 & 0.715 & 0.503 & 4.872 & 0.220 & 0.702 & 0.442 & 4.634 \\
    A.X-4.0-VL-Light & 0.277 & 0.727 & 0.482 & 3.731 & 0.238 & 0.722 & 0.461 & 3.623 \\
    HCX-SEED-Vision-3B & 0.284 & 0.735 & 0.493 & 3.776 & 0.230 & 0.718 & 0.451 & 3.671 \\
    \midrule
    
    \multicolumn{9}{c}{\textit{Open-weight large language models}} \\
    \midrule
    Qwen3-8B & 0.294 & \textbf{0.746} & 0.514 & 3.754 &  &  &  &  \\
    Qwen3-32B & 0.282 & 0.742 & 0.530 & 4.433 &  &  &  &  \\
    Gemma-2-9B & 0.288 & 0.733 & 0.518 & 4.176 &  &  &  &  \\
    Gemma-2-27B & 0.275 & 0.716 & 0.489 & 4.554 &  &  &  &  \\
    MedGemma-27B-Text & 0.288 & 0.739 & 0.546 & 4.794 &  &  &  &  \\
    A.X-4.0-Light & 0.289 & 0.730 & 0.502 & 4.212 &  &  &  &  \\
    HCX-SEED-Text-1.5B & \underline{0.295} & \underline{0.743} & 0.483 & 3.479 & & & & \\
    \bottomrule
    \end{tabular}
}
\caption{Zero-shot performance measured on \mydatabench. \textbf{Bold} and \underline{underlined} values indicate the best and second-best results, respectively. Results for closed LVLMs are averaged over three runs.}
\label{tab:zero_shot_results}
\end{table*}

\subsection{Evaluation}
\label{sec:evaluation}
We used four evaluation metrics to assess whether model-generated responses align with expert-provided answers and provide helpful information. Specifically, we adopted two traditional metrics and two LLM-as-a-judge metrics, following recent practices in medical QA~\cite{zhang-etal-2025-llmeval,he-etal-2025-astrid}. Additional details are provided in Appendix~\ref{appendix:evaluation_details}.

\paragraph{ROUGE} A reference-based metric widely used to measure lexical overlap between model-generated and reference answers~\cite{lin-2004-rouge}. We used ROUGE-L in this study.

\paragraph{BERTScore} A metric that captures semantic similarity between generated and reference answers using BERT-based contextual embeddings~\cite{Zhang2020BERTScore}, thereby complementing the lexical evaluation provided by ROUGE.

\paragraph{Factuality} A reference-based LLM-as-a-judge metric~\cite{akhtar2026ev2r} using Gemini 2.5 Flash as the judge. Following FactScore~\cite{min2023factscore}, the judge decomposes the reference and model-generated answers into atomic facts and measures their factual alignment.

\paragraph{Helpfulness} A reference-free LLM-as-a-judge metric~\cite{zhang2025longreward}. GPT-4o~\cite{hurst2024gpt} served as the judge, rating the relevance and informativeness of each model-generated answer with respect to the question on a five-point Likert scale.

\begin{figure*}[t]
    \centering
    \includegraphics[width=\textwidth]{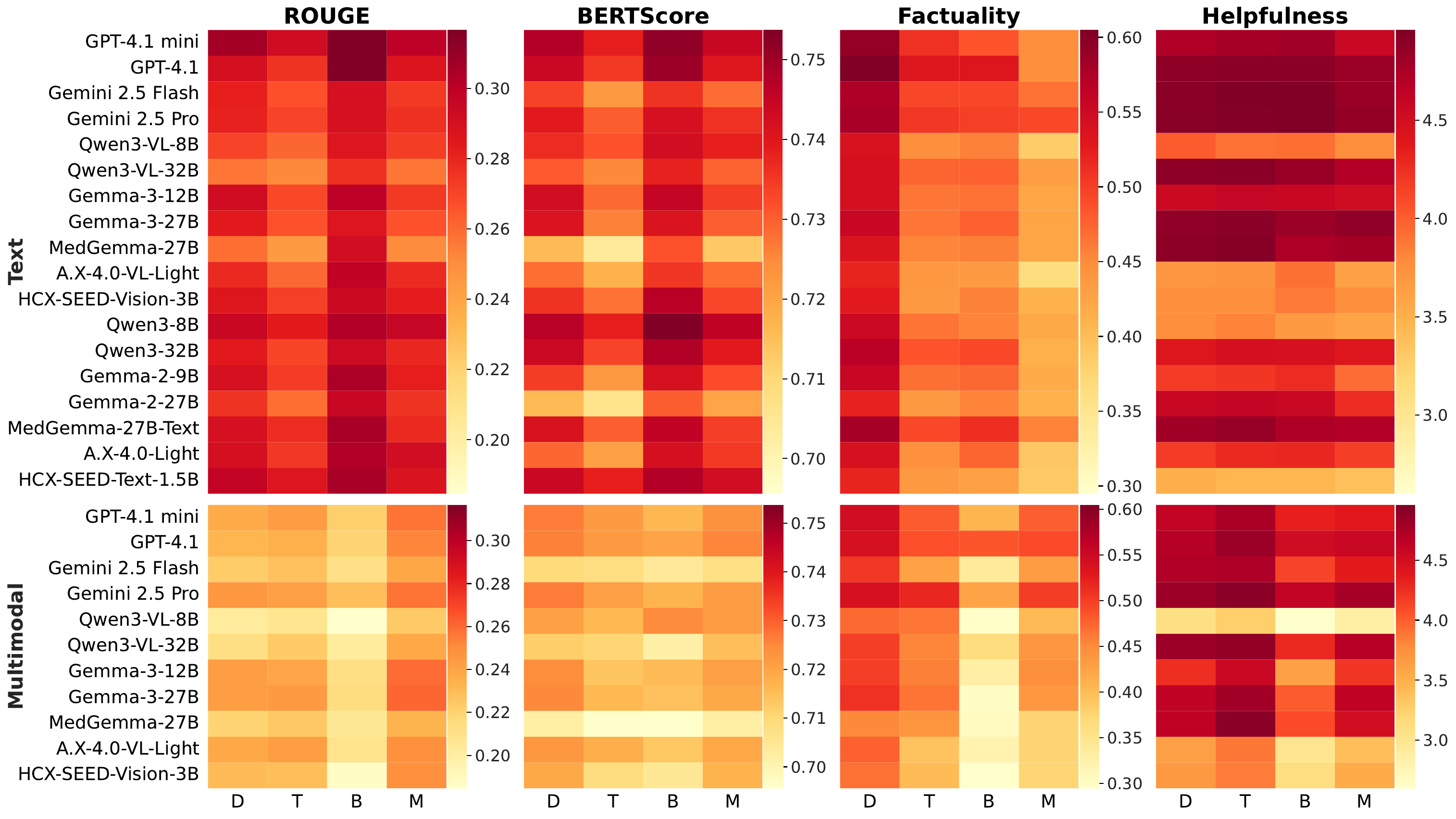}
    \caption{Performance by question type on \mydatabench. For each metric, the x-axis denotes question type: D (Diagnosis), T (Treatment), B (Basic veterinary knowledge), and M (Miscellaneous).}
    \label{fig:question_type}
\end{figure*}

\section{Benchmarking Results}

We present the benchmarking results of eighteen models on \mydatabench, focusing on veterinary knowledge and clinical reasoning. We report the zero-shot performance of the target models as our main results and then investigate the effects of RAG and SFT as alternative strategies for model adaptation. Qualitative case studies are provided in Appendix~\ref{appendix:case_study}. We additionally report benchmarking results for the English version of \mydatabench in Table~\ref{tab:zero_shot_english} and for the German, Chinese, Indonesian, and Arabic versions in Table~\ref{tab:zero_shot_multilingual}.

\subsection{Zero-Shot Performance}
\label{subsec:zero_shot}
We summarize three key findings from Table~\ref{tab:zero_shot_results}. We conducted Wilcoxon signed-rank tests to assess the statistical significance of pairwise performance differences. Claims in the text regarding performance differences were made only when the corresponding pairwise comparison was statistically significant. All findings identified below are statistically significant (\emph{p}$<$0.05).
\paragraph{Closed models achieved the highest performance in most cases.}
Closed models achieved the highest performance in most cases across both datasets. In the \textsf{Text} set, GPT-4.1 mini achieved the highest ROUGE score (0.303) and BERTScore (0.746), while Qwen3-8B achieved a comparable BERTScore. GPT-4.1 achieved the highest factuality score (0.569), and Gemini 2.5 Pro achieved the highest helpfulness score (4.923). In the \textsf{Multimodal} set, Gemini 2.5 Pro ranked highest in ROUGE (0.246) and helpfulness (4.814) and was one of the top-performing models in terms of BERTScore (0.726). GPT-4.1 and GPT-4.1 mini achieved the highest factuality scores (0.531). Some open-weight models remained competitive. In the \textsf{Text} set, MedGemma-27B-Text performed comparably to the best-performing models across metrics. Qwen3-8B achieved the highest BERTScore (0.746), but its LLM-as-a-judge scores were substantially lower than those of the top-performing models. Gemma-3-27B showed a similar trend in the \textsf{Multimodal} set, achieving competitive scores only on traditional metrics.

\paragraph{Factuality and helpfulness increased with model size, whereas ROUGE and BERTScore did not.} Within the same model family, factuality and helpfulness scores generally increased with model size, as exemplified by Qwen3-VL. There were several exceptions, including Gemma-2 for factuality. In contrast, ROUGE and BERTScore showed no consistent relationship with model size, suggesting the limitations of traditional metrics and the value of complementary LLM-as-a-judge evaluations.

\paragraph{LVLMs underperformed on the \textsf{Multimodal} set.} Across model comparisons, performance on the \textsf{Multimodal} set was consistently lower than that on the \textsf{Text} set. To examine whether differences in question-type distributions contributed to this trend, we conducted additional comparisons using only questions from the most frequent question type and sets resampled to match the question-type distribution; the same trend persisted (Appendix~\ref{appendix:additional_results}). Thus, the observed differences may reflect the greater difficulty of the multimodal questions, limitations of current LVLMs, or both, rather than differences in question-type distributions.

\begin{figure}[t]
    \centering
    \includegraphics[width=\columnwidth]{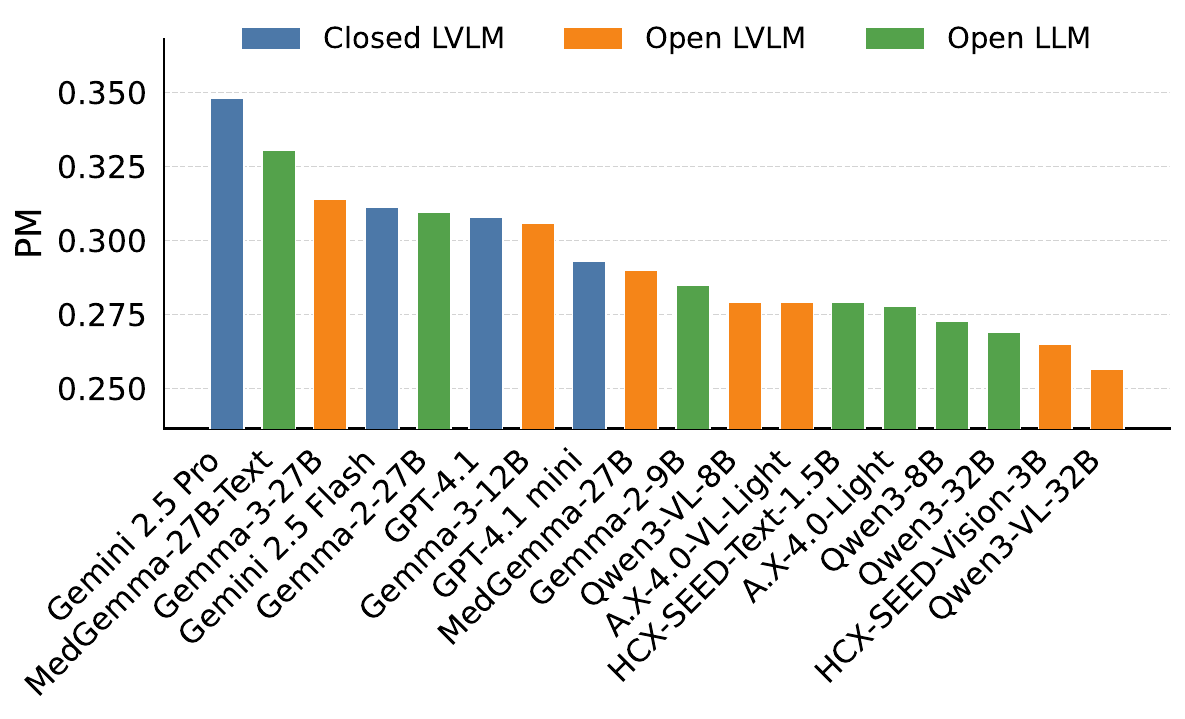}
    \caption{Clinical condition inclusion performance on the \textsf{Text} set.}
    \label{fig:clinical_condition_inclusion_performance_text}
\end{figure}

\begin{figure*}[t]
    \centering
    \includegraphics[width=\textwidth]{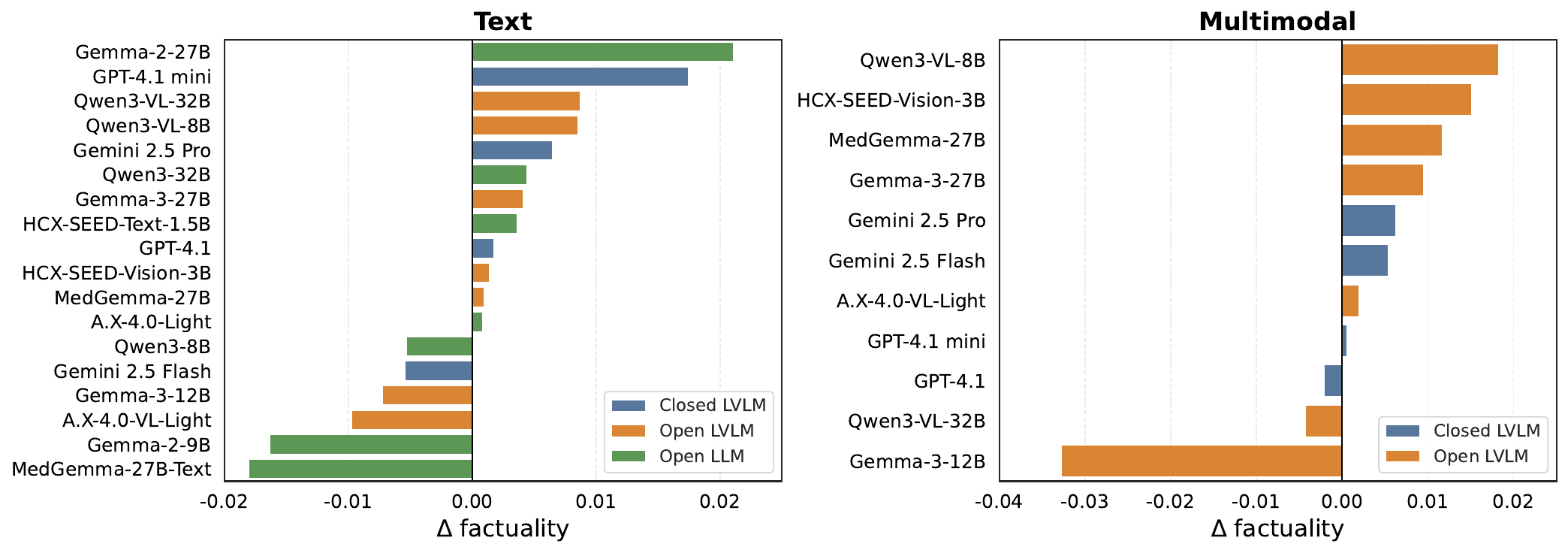}
    \caption{$\Delta$ factuality between RAG and zero-shot inference, where positive values indicate the superiority of RAG.}
    \label{fig:factuality_delta}
\end{figure*}

\subsubsection{Performance by Question Type}
\label{sec:question_type_perf}

Figure~\ref{fig:question_type} presents heatmaps illustrating model performance by question type on the \textsf{Text} and \textsf{Multimodal} sets. Using the Kruskal--Wallis test, we found statistically significant differences in scores for every metric-set combination (\emph{p}$<$0.001). 

On the \textsf{Text} set, the target models tended to achieve the best ROUGE and BERTScore performance on questions about basic veterinary knowledge. In contrast, they achieved the highest factuality scores on diagnosis questions. For helpfulness, we did not observe consistent differences across question types. When comparing model groups, closed LVLMs consistently outperformed the other models across question types, indicating greater robustness in generating helpful responses. By contrast, most open-weight models substantially underperformed in terms of helpfulness, except for Qwen3-VL-32B, MedGemma-27B, and MedGemma-27B-Text.

On the \textsf{Multimodal} set, all LVLMs achieved the highest factuality scores on diagnosis questions, consistent with the findings on the \textsf{Text} set. They achieved the highest ROUGE scores on miscellaneous questions. For BERTScore, we observed no substantial differences across question types, again consistent with the findings on the \textsf{Text} set.

\subsubsection{Clinical Condition Inclusion}
\label{sec:clinical_condition_text}

To further assess the helpfulness of model-generated responses to diagnosis questions while complementing reference-based metrics that consider the entire response, we examined whether model responses included relevant clinical conditions, such as disease names or symptoms. We used the partial match (PM) score~\cite{li2022multispanqa}, which measures overlap between model responses and ground-truth clinical conditions.

Figure~\ref{fig:clinical_condition_inclusion_performance_text} shows performance on the \textsf{Text} set, ranked by PM score. Gemini 2.5 Pro performed best, outperforming the second-best model by 0.018. Notably, MedGemma-27B-Text and Gemma-3-27B ranked second and third, outperforming the other closed models. To investigate whether the inclusion of clinical conditions was associated with performance on the four primary evaluation metrics, we further divided the diagnosis responses generated by Gemini 2.5 Pro for 694 QA pairs with non-empty ground-truth clinical conditions into matched (PM $>0$; $N=510$) and unmatched (PM $=0$; $N=184$) groups. The matched group obtained higher ROUGE (0.296 > 0.277), BERTScore (0.744 > 0.739), factuality (0.607 > 0.571), and helpfulness (4.947 > 4.929). These results suggest that the inclusion of relevant clinical conditions may be an important characteristic of accurate and helpful responses in veterinary QA.

The corresponding results for the \textsf{Multimodal} set are shown in Figure~\ref{fig:clinical_condition_inclusion_performance_multimodal} and discussed in Appendix~\ref{appendix:additional_results}.

\subsection{Retrieval-Augmented Generation}

RAG produced model-, metric, and modality-dependent effects. On the \textsf{Text} set, factuality improved for 12 of 18 models, whereas ROUGE and BERTScore decreased for 12 and 11 models, respectively; helpfulness decreased for 10 models. On the \textsf{Multimodal} set, factuality and helpfulness each improved for 8 of 11 models, while ROUGE and BERTScore decreased for 8 and 9 models, respectively. GPT-4.1 achieved the highest factuality score (0.571) and helpfulness score (4.931) on the \textsf{Text} set. For ROUGE, GPT-4.1 mini and Gemma-2-9B achieved the highest performance (0.299), while Qwen3-8B achieved the highest BERTScore (0.746). Gemini 2.5 Pro led all metrics on the \textsf{Multimodal} set. Full results are provided in Table~\ref{tab:rag_results}. 

Figure~\ref{fig:factuality_delta} shows the change in factuality, a core dimension in medical scenarios~\cite{wang2025trustworthy, wu2025medical}. On the \textsf{Text} set, Gemma-2-27B and GPT-4.1 mini showed the largest gains, whereas MedGemma-27B-Text and Gemma-2-9B showed performance degradation. On the \textsf{Multimodal} set, the largest gain was comparable to that observed on the \textsf{Text} set, whereas the largest decrease was 0.033 for Gemma-3-12B. Qualitative examples illustrate that succesful retreival of relevant information helped the model correct a zero-shot misdiagnosis (Figure~\ref{fig:rag_case2_good_case}), whereas irrelevant retrieved evidence led to an incorrect diagnoisis by causing the model to prioritize textual evidence while disregarding the input image (Figure~\ref{fig:rag_case2_bad_case}).

\subsection{Supervised Fine-Tuning}
\label{sec:sft}

SFT produced a trade-off across metrics. On the \textsf{Text} set, ROUGE and BERTScore improved for all 14 models, whereas factuality decreased for 11 and helpfulness decreased for all 14. On the \textsf{Multimodal} set, ROUGE improved for all 7 models and BERTScore improved for 6, while factuality decreased for 6 and helpfulness decreased for all 7. On the \textsf{Text} set, MedGemma-27B-Text achieved the best performance with a BERTScore of 0.780, a factuality score of 0.524, and a helpfulness score of 2.943. For ROUGE, Gemma-3-27B ranked highest at 0.366, followed by MedGemma-27B-Text at 0.362. On the \textsf{Multimodal} set, MedGemma-27B ranked highest across all metrics. Full results are provided in Table~\ref{tab:sft_results}.

Compared to the zero-shot setting, factuality improved only for Gemma-2-27B ($+0.028$) and MedGemma-27B ($+0.015$) on the \textsf{Text} set, and only for MedGemma-27B ($+0.034$) on the \textsf{Multimodal} set. To examine this trade-off, we qualitatively analyzed 100 randomly sampled responses from Qwen3-VL-32B on the \textsf{Multimodal} set, where factuality decreased after SFT (Figure~\ref{fig:sft_multimodal}). After SFT, the model responses became shorter, more generic, and more conservative. These changes improved surface-level similarity but reduced explanatory richness and coverage. This observation aligns with prior findings that a fine-tuned model may mimic response styles while remaining factually unreliable~\cite{gudibande2024the}.

\section{Conclusion}
This study introduces \mydata, a Korean QA benchmark designed to evaluate the veterinary knowledge and clinical reasoning capabilities of LLMs and LVLMs. \mydata comprises both text-only and image-based questions, along with answers provided by expert veterinarians, based on real-world cases involving dogs and cats. \mydatabench, the test set of \mydata, additionally includes annotations for question type and clinical condition. To the best of our knowledge, \mydata is the first long-form QA resource for veterinary medicine. To support the broader NLP community, we release translated versions of \mydatabench in English, German, Chinese, Indonesian, and Arabic.

We conduct benchmarking experiments for eighteen models categorized as closed LVLMs, open-weight LVLMs, and open-weight LLMs. Two traditional metrics and two LLM-as-a-judge metrics are used to measure the alignment between model responses and expert-annotated references and to assess the helpfulness of the responses. The results reveal three key findings. First, closed models generally outperformed open-weight models across the four metrics, with larger gaps observed for the LLM-as-a-judge metrics of factuality and helpfulness. Second, all models showed consistent performance degradation on image-based questions. Third, alternative model adaptation strategies, such as RAG and SFT, yielded inconsistent improvements on both the \textsf{Text} and \textsf{Multimodal} sets. Overall, these benchmarking results highlight the limitations of current models and the need for more effective adaptation methods to develop clinically reliable AI systems for veterinary care.

\section*{Limitations}

\paragraph{Dataset Coverage} Our dataset focuses on two companion animals, dogs and cats, which account for a substantial proportion of real-world pet-related queries. Given Naver Knowledge iN's significant user base, the dataset likely encompasses a wide range of clinical queries about these target animals. As identified in Section~\ref{subsec:descriptive_analysis}, \mydatabench covers 352 clinical conditions across 4,000 QA pairs, further demonstrating its diversity. Future work could extend the dataset to a broader range of species and investigate the generalizability of our findings across languages. We hope that the data collection, preprocessing, and annotation protocols documented in this study can facilitate the development of resources covering a wider range of species and languages.

\paragraph{Evaluation Metrics} Results measured by reference-based metrics may be misleading when the reference answers are incorrect. To assess the reliability of the expert-provided answers, we verified the medical accuracy of responses collected from the online QA platform through manual validation by the fourth author, a veterinary expert holding a Ph.D. (Section~\ref{sec:annotation}). These results suggest that expert-based quality checks should be incorporated into future efforts to construct similar resources. 

\section*{Ethical Considerations}
This study was approved by the Institutional Review Board at Soongsil University (SSU-202604-HR-805-1). 
\paragraph{Copyright and Privacy Issues}
To comply with the restrictions specified in the Terms of Service of Naver Knowledge iN, we release the dataset for research purposes only under the CC BY-NC-ND 4.0 license, which restricts commercial use and the distribution of derivative works.
We manually confirmed that \mydatabench do not contain personally identifiable information. 
\paragraph{Annotator Information}
Label annotation for \mydatabench (Section~\ref{sec:annotation}) was conducted by six students majoring in AI recruited from two institutions: four master's students and two undergraduate students. The annotation task primarily involved reading text instances and assigning labels. Therefore, the annotation task was considered minimal risk, with no anticipated physical or psychological harm to participants. In compliance with local wage regulations, the annotators were compensated at approximately USD 7 per hour. The quality check described in the same section was performed by a veterinary expert holding a Ph.D. Screenshots of the annotation interfaces are available in Figures~\ref{fig:annotator_interface} and \ref{fig:expert_validation_interface}.

\paragraph{AI Assistant Use}
We used AI-assisted language-editing tools, primarily ChatGPT, exclusively for checking grammar and improving readability.

\section*{Acknowledgements}
This research was supported by the IITP (Institute of Information \& Communications Technology Planning \& Evaluation), funded by the Korea government (MSIT) (IITP-2026-RS-2022-00156360, IITP-2026-RS-2024-00430997, IITP-2026-RS-2020-II201602). This work was also supported by a grant from the Korea Health Technology R\&D Project through the Korea Health Industry Development Institute (KHIDI), funded by the Ministry of Health \& Welfare, the Republic of Korea (Grant No. HI22C0646). KP and SP are the corresponding authors.

\bibliography{main}

\appendix

\setcounter{table}{0}
\setcounter{figure}{0}
\renewcommand\thefigure{A\arabic{figure}} 
\renewcommand\thetable{A\arabic{table}}

\section{Artifact Usage}
All artifacts, including datasets, reference materials, and evaluation packages, were used strictly for non-commercial research and evaluation purposes. Below, we document the scope and use of our artifacts and confirm that their use is consistent with their intended purposes.

\subsection{Source Data}

\paragraph{Veterinary QA Data}
\mydata consists of Korean veterinary long-form QA data collected from publicly accessible Naver Knowledge iN pages. The data focus on clinical consultations for companion animals, specifying dogs and cats.
Before data collection, we reviewed the platform's terms of service and posting policy\footnote{\url{https://kin.naver.com/common/guide.naver?query=p3}}. According to these policies, the copyright of user-generated posts belongs to the original authors, and the platform permits limited use for service operation, search, and research-related purposes\footnote{\url{https://policy.naver.com/policy/service.html}}. In accordance with these policies and prior work~\cite{choi2025chatgpt}, we de-identified the collected data by removing personally identifiable information and do not redistribute the original raw posts. 

\paragraph{Retrieval Corpus}
The retrieval corpus was constructed from an English-language veterinary reference book intended for professional veterinary education and commonly used in clinical practice.\footnote{Legal constraints prevent us from disclosing the title.} The book was used solely as an external knowledge source for retrieval-augmented evaluation and was not used for model training.

\paragraph{Clinical Condition}
To support clinical condition annotation, we additionally used Korean veterinary and medical reference resources, including the official animal disease classification system~\cite{mafra-animal-disease}, disease encyclopedias~\cite{asan-medical-center}, and disease information resources~\cite{kpic-disease-info}. These resources were used only internally for annotation and terminology normalization. We do not redistribute their original webpage content or any source text as part of the dataset or research artifacts.

\subsection{License}
We report the explicitly stated licenses or terms associated with the models used in this work.

The GPT model series were used in compliance with the OpenAI Terms of Use~\cite{openaitou}. Gemini models~\cite{comanici2025gemini} were used in compliance with the Google APIs Terms of Service~\cite{geminiterm}. Qwen3~\cite{qwen3embedding,bai2025qwen3,qwen3technicalreport} and A.X models~\cite{SKTAdotX4Light,SKTAdotX4VLLight} are available under the Apache License 2.0. Gemma models~\cite{team2024gemma,gemma3} are available under the Gemma Terms of Use~\cite{gemmaterm}. MedGemma models are available under the Health AI Developer Foundations Terms of Use~\cite{medgemmaterm}. HCX models~\cite{hcx_seed_vision_3b,hcx_seed_text_1_5b} are available under the HyperCLOVA X SEED Model License Agreement~\cite{hcxlicense}. EXAONE-3.5-32B~\cite{an2024exaone} is available under the EXAONE AI Model License Agreement 1.1 - NC~\cite{exaonelicense}. KLUE-RoBERTa-base~\cite{park2021klue} is available under the CC BY-SA 4.0 license.

\section{Dataset Details}
\label{appendix:dataset_details}

\paragraph{Raw Data Collection}
We collected posts from Naver Knowledge iN through keyword- and profile-based strategies. First, we searched for dog- and cat-related keywords appearing in questions and answers. Second, we collected posts from verified experts' profile pages, ensuring that each post contained at least one expert answer. Each post was crawled using Selenium\footnote{\url{https://www.selenium.dev/}}, and its HTML was parsed with BeautifulSoup\footnote{\url{https://www.crummy.com/software/BeautifulSoup/}}. We extracted the question title and body, all answers, answer selection status, board information, hashtags, and expert verification status.

\paragraph{Animal Type Classification}
To distinguish dog- and cat-related posts, we used a two-stage pipeline combining hashtag-based labeling and text classification. Posts matching hashtags from only one of the two lists containing the 100 most frequent dog- and cat-related hashtags were labeled accordingly, with the hashtags serving as weak labels~\cite{mohammad-2012-emotional}. The remaining posts were classified as \textit{dog}, \textit{cat}, or \textit{neutral} using two fine-tuned KLUE-RoBERTa-base models, each trained separately on posts with and without images. After removing URLs, we used the question title, body, and selected answer as input. Neutral examples were drawn from seven non-target animal boards, and posts classified as \textit{neutral} were excluded from the dataset.

\paragraph{Training and Validation Set Statistics}

\begin{table}[t]
\centering
\small
\setlength{\tabcolsep}{3pt}
\begin{tabular}{lrrrr}
\toprule
\textbf{Metric} & \multicolumn{2}{c}{\textbf{Train}} & \multicolumn{2}{c}{\textbf{Validation}} \\
 & \textsf{Text} & \textsf{Multimodal} & \textsf{Text} & \textsf{Multimodal} \\
\midrule
\textbf{Target species} & & & & \\
Dog & 5,098 & 3,871 & 1,675 & 1,595 \\
Cat & 978 & 880 & 325 & 405 \\
\midrule
\textbf{Text tokens} & & & & \\
\emph{Questions} & & & & \\
- Max. & 971 & 1016 & 1339 & 1128 \\
- Mean. & 145.3 & 107.9 & 142.6 & 110.7 \\
- Min. & 10 & 6 & 10 & 9 \\
\emph{Answers} & & & & \\
- Max. & 544 & 785 & 396 & 761 \\
- Mean. & 120.9 & 139.0 & 119.2 & 139.9 \\
- Min. & 6 & 7 & 11 & 17 \\
\bottomrule
\end{tabular}
\caption{Descriptive statistics of \mydata for the train and validation sets.}
\label{tab:petqa_statistics_train_val}
\end{table}

Table~\ref{tab:petqa_statistics_train_val} provides descriptive statistics for the training and validation sets.

\section{Evaluation Metrics}
\label{appendix:evaluation_details}

Following Section~\ref{sec:evaluation}, we describe the implementation details and protocols for all metrics. For LLM-as-a-judge evaluations, we used a fixed random seed of 42 and a temperature of 0 to ensure reproducibility and consistency across evaluations. All LLM-based evaluations were conducted using the batch API, and the evaluation prompts are provided in Appendix~\ref{appendix:prompts}.

\paragraph{ROUGE}
We computed ROUGE~\cite{lin-2004-rouge} using the \texttt{rouge} Python package (v1.0.1). To better reflect Korean morphology, we tokenized both reference answers. We generated answers at the morpheme level using KoNLPy (v0.6.0)~\cite{park2014konlpy} with the Okt tokenizer, and computed ROUGE on the resulting whitespace-joined morpheme sequences. We reported the mean F1 score over all examples.

\paragraph{BERTScore}
We computed BERTScore~\cite{Zhang2020BERTScore} using the \texttt{bert-score} package (v0.3.13). We used the official Python API with the default multilingual checkpoint (\texttt{bert-base-multilingual-cased}), and reported the mean F1 score over all examples.

\paragraph{Factuality}

We evaluated factuality using a reference-based metric~\cite{akhtar2026ev2r}, with Gemini 2.5 Flash as the judge model due to its comparable agreement with human judgments and greater cost efficiency, as shown in Table~\ref{tab:judge_human_correlation_factuality}. Following FactScore~\cite{min2023factscore}, the judge decomposed the reference answer ($y$) and model output ($\hat{y}$) into atomic fact sets $A_y$ and $A_{\hat{y}}$, respectively. We computed precision ($P$) as the proportion of generated facts supported by the reference and recall ($R$) as the proportion of reference facts covered by the model output:
\begin{align}
P &= \frac{1}{|A_{\hat{y}}|} \sum_{a_{\hat{y}} \in A_{\hat{y}}} \mathbb{I}[a_{\hat{y}} \text{ is supported by } y] \\
R &= \frac{1}{|A_y|} \sum_{a_y \in A_y} \mathbb{I}[a_y \text{ is supported by } \hat{y}] 
\end{align}
where $\mathbb{I}[\cdot]$ denotes the indicator function. We reported the mean F1 score as the primary factuality metric, capturing both the factual accuracy and completeness of generated answers. We used a thinking budget of 0 and a maximum output length of 8,192 tokens. The prompt is shown in Figure~\ref{fig:factuality_judge_prompt_en}.

\paragraph{Helpfulness}
We evaluated helpfulness using a reference-free metric. Following~\citet{zhang2025longreward}, we provided the judge model with detailed scoring criteria and examples for each score level. Given a question and model response, the judge model assigns a score and generate a brief rationale. The prompt is shown in Figure~\ref{fig:helpfulness_judge_prompt_en}. 

To select the judge model, we randomly sampled 100 responses from each set and manually annotated them on a five-point Likert scale. We reported the mean helpfulness score and used a maximum output length of 1,024 tokens. As shown in Table~\ref{tab:judge_human_correlation_helpfulness}, GPT-4o exhibited the highest overall correlation and was therefore selected. It matched the human scores exactly for 58 samples and differed by at most one point for an additional 38 samples. Although the modest correlation does not necessarily imply that system-level comparisons are unreliable, as discussed in prior work~\cite{gera-etal-2025-justrank}, the helpfulness results should be interpreted with caution and further validated in future work.

\paragraph{Partial Match}
Following \citet{li2022multispanqa}, we evaluated predicted and gold clinical condition spans using partial matching. We used EXAONE-3.5-32B to extract predicted conditions from model responses. Let $p_i$ and $g_j$ denote the $i$-th predicted and $j$-th gold condition, respectively. To account for minor span variations, we define their retrieval and relevance overlap scores using the longest common substring (LCS):
\begin{equation}
    s^{ret}_{ij} = \frac{len(LCS(p_i,g_j))}{len(p_i)}
\end{equation}
\begin{equation}
    s^{rel}_{ij} = \frac{len(LCS(p_i,g_j))}{len(g_j)}
\end{equation}
Given $n$ predicted and $m$ gold conditions, partial precision and recall are computed as:
\begin{equation}
    Precision=\frac{\sum_{i=1}^{n}\max_{j \in [1,m]}(s^{ret}_{ij})}{n}
\end{equation}
\begin{equation}
    Recall=\frac{\sum_{j=1}^{m}\max_{i \in [1,n]}(s^{rel}_{ij})}{m}
\end{equation}
We reported their harmonic mean as the micro-averaged F1 score.

\begin{table}[t]
\centering
\small
\begin{tabular}{lccc}
\toprule
\textbf{Judge Model} & \textbf{Pearson} & \textbf{Spearman} & \textbf{Kendall} \\
\midrule
GPT-4o           & \textbf{0.2255}  & \textbf{0.2239}  & \textbf{0.2144} \\
GPT-5            & 0.1544  & 0.2218  & 0.2097 \\
Gemini 2.5 Pro   & 0.0967  & 0.1231  & 0.1149 \\
EXAONE-3.5-32B   & -0.1097 & -0.0781 & -0.0771 \\
\bottomrule
\end{tabular}
\caption{Correlation between human evaluation scores and LLM-as-a-judge scores for helpfulness.}
\label{tab:judge_human_correlation_helpfulness}
\end{table}

\begin{table}[t]
\centering
\small
\begin{tabular}{lcc}
\toprule
\textbf{Judge Model} & \textbf{Pearson} & \textbf{Spearman} \\
\midrule
Gemini 2.5 Pro   & \textbf{0.4087} & \textbf{0.3922} \\
Gemini 2.5 Flash & 0.4065          & 0.3784          \\
GPT-4o-mini      & 0.1514          & 0.0995          \\
\bottomrule
\end{tabular}
\caption{Correlations between human evaluation scores and LLM-as-a-judge factuality scores for the factuality.}
\label{tab:judge_human_correlation_factuality}
\end{table}

\section{Experimental Setups}
\label{appendix:experimental_setups}

\paragraph{Computing Environment}
Experiments were conducted on two separate computing environments. The first system was equipped with three NVIDIA RTX A6000 GPUs (48 GB of VRAM each) and 128 GB of system memory, while the second system consisted of a single NVIDIA H200 GPU (141 GB of VRAM) and 2 TB of system memory. All experiments were performed using Python 3.12.0, PyTorch 2.8.0, Transformers 4.57.3, and vLLM 0.11.0.

\paragraph{Models}
Table~\ref{tab:models_info} summarizes the model IDs and parameter sizes of the models used in our experiments.
Following practices adopted in previous studies~\cite{lee2025koblex}, we disabled reasoning (or thinking) modes for models that support explicit reasoning functionalities (e.g., the Qwen3 and Gemini 2.5 families).

\begin{table*}[t]
\centering
\small
\begin{tabularx}{\textwidth}{lXc}
\toprule
\textbf{Model} & \textbf{Model ID} & \textbf{Params}\\
\midrule
\multicolumn{3}{c}{\textit{Closed large vision-language models}} \\
\midrule
GPT-4.1 mini & \texttt{gpt-4.1-mini-2025-04-14} & -- \\
GPT-4.1 & \texttt{gpt-4.1-2025-04-14} & -- \\
Gemini 2.5 Flash & \texttt{gemini-2.5-flash} & -- \\
Gemini 2.5 Pro & \texttt{gemini-2.5-pro} & -- \\
\midrule
\multicolumn{3}{c}{\textit{Open-weight large vision-language models}} \\
\midrule
Qwen3-VL-8B & \texttt{Qwen/Qwen3-VL-8B-Instruct} & 8B \\
Qwen3-VL-32B & \texttt{Qwen/Qwen3-VL-32B-Instruct} & 32B \\
Gemma-3-12B & \texttt{google/gemma-3-12b-it} & 12B \\
Gemma-3-27B & \texttt{google/gemma-3-27b-it} & 27B \\
MedGemma-27B & \texttt{google/medgemma-27b-it} & 27B \\
A.X-4.0-VL-Light & \texttt{skt/A.X-4.0-VL-Light} & 8B \\
HCX-SEED-Vision-3B & \texttt{naver-hyperclovax/HyperCLOVAX-SEED-Vision-Instruct-3B} & 3B \\
\midrule
\multicolumn{3}{c}{\textit{Open-weight large language models}} \\
\midrule
Qwen3-8B & \texttt{Qwen/Qwen3-8B} & 8B \\
Qwen3-32B & \texttt{Qwen/Qwen3-32B} & 32B \\
Gemma-2-9B & \texttt{google/gemma-2-9b-it} & 9B \\
Gemma-2-27B & \texttt{google/gemma-2-27b-it} & 27B \\
MedGemma-27B-Text & \texttt{google/medgemma-27b-text-it} & 27B \\
A.X-4.0-Light & \texttt{skt/A.X-4.0-Light} & 7B \\
HCX-SEED-Text-1.5B & \texttt{naver-hyperclovax/HyperCLOVAX-SEED-Text-Instruct-1.5B} & 1.5B \\
\bottomrule
\end{tabularx}
\caption{Models used in experiments, along with their corresponding model identifiers and parameter sizes.}
\label{tab:models_info}
\end{table*}

\paragraph{Inference and Training Settings}
For all open-weight models, inference was conducted using vLLM~\cite{kwon2023efficient} with a fixed random seed of 42, a temperature of 0, a repetition penalty of 1.05, and a maximum generation length of 512 tokens. For RAG, the top six passages were retrieved from a FAISS index implemented with LangChain using embeddings generated by Qwen3-Embedding-8B. Documents were divided into 1,024-token chunks with an overlap of 100 tokens.

For parameter-efficient fine-tuning, we used LoRA~\cite{hu2022lora} with a rank ($r$) of 16, a dropout rate of 0.05, and an $\alpha$ value of 32. Models were trained for two epochs using AdamW with a learning rate of $5\times10^{-5}$, a weight decay of 0.01, a warmup ratio of 0.1, four gradient accumulation steps, and a batch size of 8. Training and inference required approximately 15 and 12 GPU hours, respectively.

\begin{figure}[t]
\centering
\includegraphics[width=\columnwidth]{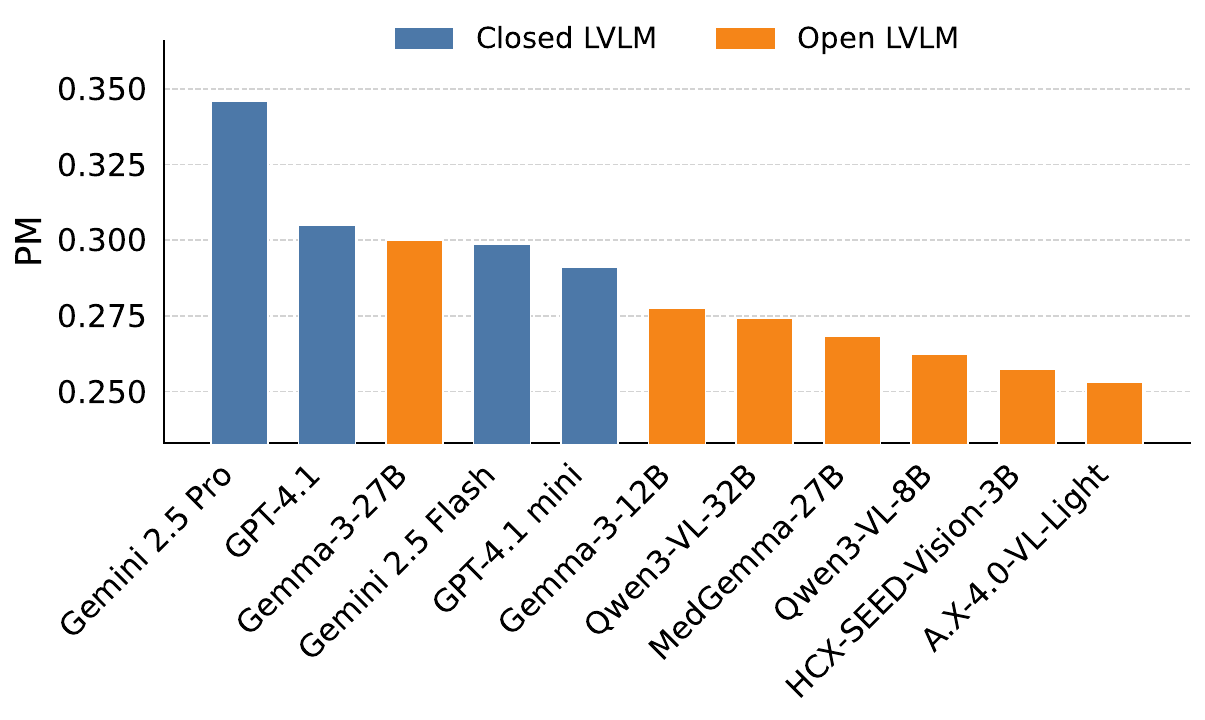}
\caption{Clinical condition inclusion performance on the \textsf{Multimodal} set.}
\label{fig:clinical_condition_inclusion_performance_multimodal}
\end{figure}

\section{Case Study}
\label{appendix:case_study}

\begin{figure*}[t]
    \centering
    \noindent
    \begin{minipage}[t]{1.0\textwidth}
        \begin{tcolorbox}[title={\small \textbf{Question}}]
            \begin{minipage}[t]{0.72\textwidth}
                My dog seemed to be in heat, so I put a diaper on her, and this came out on it. What is this?
            \end{minipage}
            \hfill
            \begin{minipage}[t]{0.22\textwidth}
                \vspace{0pt}
                \centering
                \includegraphics[width=\linewidth]{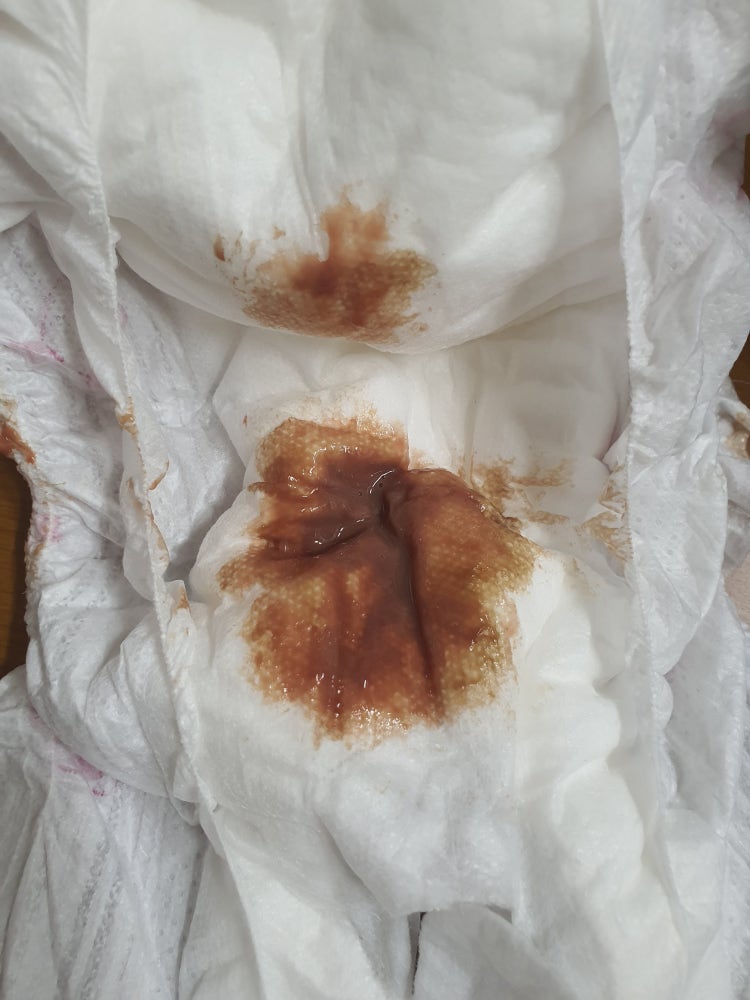}
            \end{minipage}
        \end{tcolorbox}
    \end{minipage}
        
    \begin{minipage}[t]{1.0\textwidth}
        \begin{tcolorbox}[title={\small \textbf{Reference Answer}}]
            It is highly likely to be \textbf{pyometra}. Common symptoms of pyometra include polydipsia, polyuria, vomiting, loss of appetite, abdominal distension, and pus discharge from the genital area. An accurate diagnosis should be confirmed at a veterinary clinic using ultrasonography.
        \end{tcolorbox}
    \end{minipage}
    \begin{minipage}[t]{1.0\textwidth}
        \begin{tcolorbox}[title={\small \textbf{Retrieved Documents}}]
             \textbf{Pyometra}, haemometra, mucometra - Tumour ... \\
             ... \\
             Discharge (vulvitis/vaginitis, pyometra, haemometra, mucometra) ... \\
             ... \\
             Principal Symptoms \\
             Polydipsia, Polyuria, Vaginal discharge, Behavioral changes, ...
        \end{tcolorbox}
    \end{minipage}
    
    \begin{minipage}[t]{1.0\textwidth}
        \begin{tcolorbox}[title={\small \textbf{Gemma-3-12B}}]
            \textbf{Zero-shot Output}\\
                What is shown in the picture appears to be \textbf{menstrual blood} from the dog. A dog’s discharge during the heat cycle usually changes from red to brown and may appear in clumps. The estrous cycle in dogs typically lasts between 21 and 35 days, and bleeding may continue for 2 to 10 days. However, if the amount of bleeding is excessive or accompanied by other symptoms, consulting a veterinarian is recommended.
            \tcbline
            \textbf{RAG Output}\\
                The substance found on the dog’s diaper appears to be blood. According to Documents 2 and 6, vaginal discharge may indicate uterine conditions such as \textbf{pyometra}, haemometra, or mucometra, or conditions such as vulvitis/vaginitis. In addition, behavioral changes, polyuria, and polydipsia may also occur (Document 5). It is important to have the dog examined by a veterinarian to determine the exact cause and receive appropriate treatment.
            \tcbline
            \textbf{Analysis} \\
                Unlike the zero-shot response, which misinterpreted the discharge as a normal menstrual cycle, the RAG response correctly identified the possibility of uterine conditions, such as pyometra, from the retrieved documents, demonstrating cross-modal reasoning by grounding the clinical image and the retrieved documents to produce an accurate answer.
        \end{tcolorbox}
    \end{minipage}
\caption{Qualitative example in which retrieval-augmented generation (RAG) produces an accurate response by jointly reasoning over the clinical image and retrieved documents.}
\label{fig:rag_case2_good_case}
\end{figure*}

\begin{figure*}[t]
    \centering
    \noindent
    \begin{minipage}[t]{1.0\textwidth}
        \begin{tcolorbox}[title={\small \textbf{Question}}]
            \begin{minipage}[t]{0.72\textwidth}
                 I am wondering what this pimple-like bump on the back of my 3-year-old Chihuahua’s neck could be. Is it okay to leave it untreated?
            \end{minipage}
            \hfill
            \begin{minipage}[t]{0.22\textwidth}
                \vspace{0pt}
                \centering
                \includegraphics[width=\linewidth]{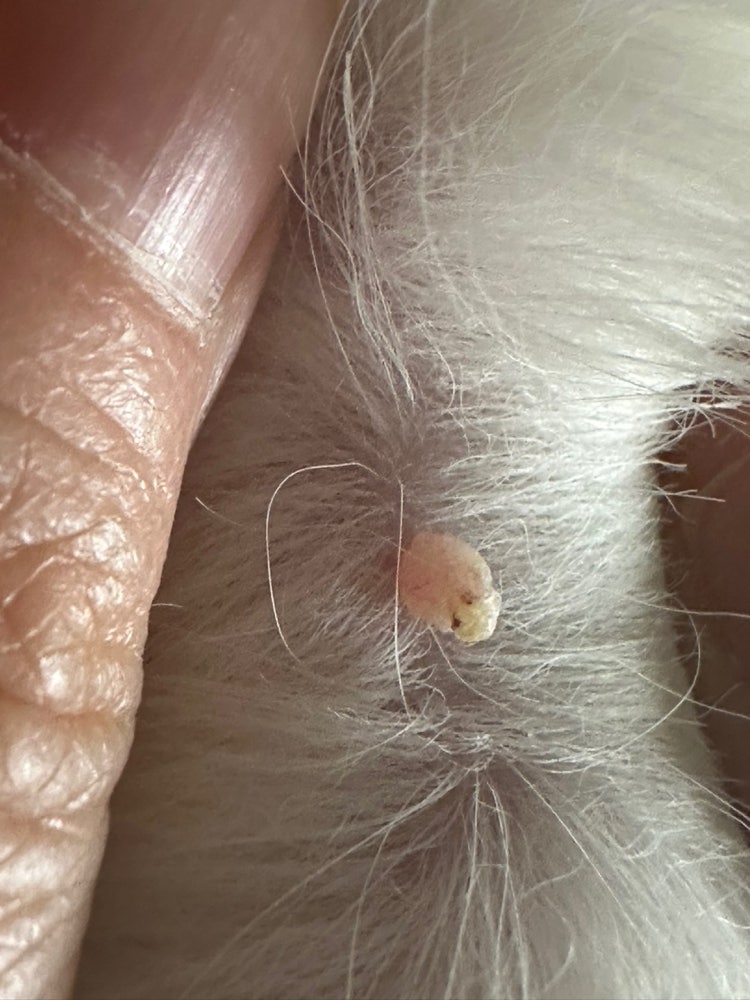}
            \end{minipage}
        \end{tcolorbox}
    \end{minipage}
    
    \begin{minipage}[t]{1.0\textwidth}
        \begin{tcolorbox}[title={\small \textbf{Reference Answer}}]
            Based on the image, it is highly likely to be a \textbf{skin tumor}, such as a papilloma, and biopsy as well as surgical treatment may be necessary. If left untreated, it may increase in size and worsen with inflammation and bleeding.
        \end{tcolorbox}
    \end{minipage}
    \begin{minipage}[t]{1.0\textwidth}
        \begin{tcolorbox}[title={\small \textbf{Retrieved Documents}}]
            \textbf{Retention cyst of the salivary gland} (ranula) \\
            Structures of the throat/neck - Retention cyst of the salivary gland (cyst on the neck/meliceris) ...
        \end{tcolorbox}
    \end{minipage}
    
    \begin{minipage}[t]{1.0\textwidth}
        \begin{tcolorbox}[title={\small \textbf{Gemma-3-12B}}]
            \textbf{Zero-shot Output}\\
            The bump shown in the image may be a \textbf{skin tumor}. If left untreated, it could grow larger or become inflamed, so it is recommended that a veterinarian examine it. Visit a veterinary clinic for an accurate diagnosis and appropriate treatment.
            \tcbline
            \textbf{RAG Output}\\
            The pimple-like lesion on the back of the 3-year-old Chihuahua’s neck may be a \textbf{retention cyst of the salivary gland}. This is also referred to as meliceris. Whether it is safe to leave untreated depends on an accurate diagnosis, so it is recommended to consult a veterinarian for an examination.
            \tcbline
            \textbf{Analysis} \\
                In the zero-shot response, the model directly interprets the lesion's appearance and correctly identifies it as a skin tumor. In contrast, RAG biases the model toward the retrieved text, which mentions a salivary gland retention cyst, causing it to prioritize lexical overlap with the question over visual evidence. As a result, the alignment between visual and textual cues breaks down, leading to an incorrect diagnosis.
        \end{tcolorbox}
    \end{minipage}
\caption{Qualitative example showing that retrieval-augmented generation (RAG) prioritizes retrieved text over visual context, leading to an incorrect diagnosis.}
\label{fig:rag_case2_bad_case}
\end{figure*}

\paragraph{RAG}
Figures~\ref{fig:rag_case2_good_case} and~\ref{fig:rag_case2_bad_case} compare model responses under the zero-shot and RAG settings using representative successful and failed cases, respectively. 

\begin{figure*}[t]
    \centering
    \noindent
    \begin{minipage}[t]{1.0\textwidth}
        \begin{tcolorbox}[title={\small \textbf{Question}}]
            \begin{minipage}[t]{0.72\textwidth}
                 A dog had a mild seizure in February and underwent X-rays and blood tests. No disk herniation was found, patellar luxation was present, and the liver enzymes were slightly elevated. The dog had another seizure today, so we are considering an MRI. The seizure lasts about 1-2 minutes, and the dog has no problems walking or running. Does everything look normal in the photos?
            \end{minipage}
            \hfill
            \begin{minipage}[t]{0.22\textwidth}
                \vspace{0pt}
                \centering
                \includegraphics[width=\linewidth]{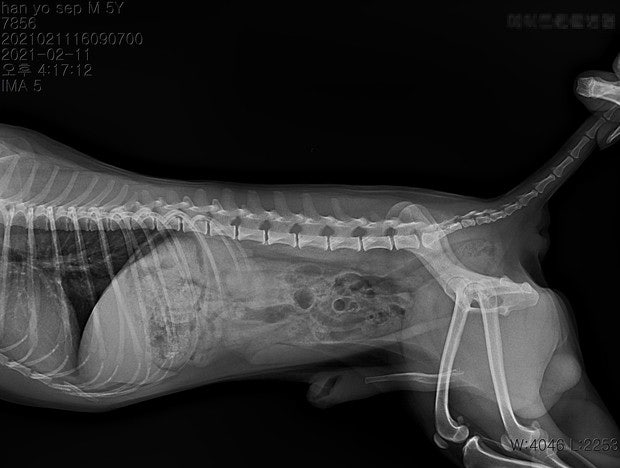}
            \end{minipage}
        \end{tcolorbox}
    \end{minipage}
    \begin{minipage}[t]{1.0\textwidth}
        \begin{tcolorbox}[title={\small \textbf{Reference Answer}}]
            Based on the submitted materials, the current condition appears to be \textbf{completely normal}. However, since the tests were performed 4-5 months ago, a follow-up examination may be necessary. Even if the previous test results were normal, it does not mean that the current condition is entirely flawless. A repeat examination is recommended, and even if acupuncture treatment is pursued, an MRI should still be performed. It is important to identify the lesion and proceed with treatment accordingly; treatment without objective evidence would be reckless. After the MRI, it would be advisable to choose rehabilitation therapy or acupuncture. I hope the problem can be resolved through prompt and accurate diagnosis and effective treatment.
        \end{tcolorbox}
    \end{minipage}
    
    \begin{minipage}[t]{1.0\textwidth}
        \begin{tcolorbox}[title={\small \textbf{Qwen3-VL-32B}}]
            \textbf{Zero-shot Output}\\
                On the X-ray image, \textbf{no spinal disk herniation or clear skeletal abnormalities are observed}. Patellar luxation can be detected on X-rays, but no definite signs of luxation are visible in this image, and further evaluation, such as MRI, may be necessary. The elevated liver enzymes and seizure symptoms raise suspicion of a neurological cause, and an MRI can more accurately assess abnormalities in the spinal cord or brain. Based on the X-ray alone, the cause of the seizures cannot be determined, and an MRI is recommended. There is no immediate skeletal abnormality visible in the photo, but additional testing is needed for neurological evaluation.
            \tcbline
            \textbf{SFT Output}\\
                The current radiograph has limited value for diagnosing a disk problem. For accurate diagnosis, the dog must be properly restrained during radiography to prevent movement. If radiography is difficult, MRI should be considered.
            \tcbline
            \textbf{Analysis} \\
                The SFT output is more general and conservative than the zero-shot response. Rather than providing a concrete visual assessment of the radiograph and discussing possible abnormalities, it offers only a cautious, generic response. Although this framing may reduce the risk of overinterpretation, it does not adequately address the user's request for an image-based assessment.
        \end{tcolorbox}
    \end{minipage}
\caption{Qualitative example showing that supervised fine-tuning (SFT) produces a more generic and conservative response with reduced image-grounded reasoning.}
\label{fig:sft_multimodal}
\end{figure*}

\paragraph{SFT}
Figure~\ref{fig:sft_multimodal} compares model responses under the zero-shot and SFT settings and analyzes their differences.

\section{Supplementary Results}
\label{appendix:additional_results}

\paragraph{Clinical Condition Inclusion}

In Section~\ref{sec:clinical_condition_text}, we assessed whether model responses included clinical conditions for the samples in the \textsf{Text} set using the partial match (PM) score. Here, we present the corresponding results for the \textsf{Multimodal} set, as shown in Figure~\ref{fig:clinical_condition_inclusion_performance_multimodal}. Results indicated that Gemini 2.5 Pro performed best, followed by GPT-4.1 and Gemma-3-27B. An analysis of 958 diagnosis responses generated by Gemini 2.5 Pro with non-empty ground-truth clinical conditions revealed a similar trend across the four evaluation metrics. The matched group obtained higher ROUGE (0.259$>$0.241), BERTScore (0.731$>$0.723), factuality (0.574$>$0.484), and helpfulness (4.836$>$4.764). 

\begin{table*}[t]
\centering
\small
\setlength{\tabcolsep}{3.5pt}
\begin{tabular}{lcccccc}
\toprule
Model
& \multicolumn{3}{c}{\textsf{Text}}
& \multicolumn{3}{c}{\textsf{Multimodal}} \\
\cmidrule(lr){2-4} \cmidrule(lr){5-7}
& P
& R
& F1
& P
& R
& F1 \\
\midrule

\multicolumn{7}{c}{\textit{Closed large vision-language models}} \\
\midrule
GPT-4.1 mini       & 0.528 & 0.647 & 0.551 & 0.562 & 0.572 & 0.531 \\
GPT-4.1            & 0.535 & 0.678 & 0.569 & 0.538 & 0.590 & 0.531 \\
Gemini 2.5 Flash   & 0.493 & 0.682 & 0.541 & 0.479 & 0.586 & 0.495 \\
Gemini 2.5 Pro     & 0.502 & 0.679 & 0.550 & 0.512 & 0.625 & 0.524 \\

\midrule
\multicolumn{7}{c}{\textit{Open-weight large vision-language models}} \\
\midrule
Qwen3-VL-8B        & 0.503 & 0.562 & 0.499 & 0.589 & 0.435 & 0.461 \\
Qwen3-VL-32B       & 0.460 & 0.659 & 0.513 & 0.468 & 0.584 & 0.489 \\
Gemma-3-12B        & 0.489 & 0.609 & 0.510 & 0.521 & 0.523 & 0.490 \\
Gemma-3-27B        & 0.475 & 0.646 & 0.519 & 0.497 & 0.564 & 0.496 \\
MedGemma-27B       & 0.450 & 0.662 & 0.503 & 0.422 & 0.546 & 0.442 \\
A.X-4.0-VL-Light   & 0.487 & 0.547 & 0.482 & 0.523 & 0.469 & 0.461 \\
HCX-SEED-Vision-3B & 0.507 & 0.552 & 0.493 & 0.530 & 0.455 & 0.451 \\

\midrule
\multicolumn{7}{c}{\textit{Open-weight large language models}} \\
\midrule
Qwen3-8B           & 0.536 & 0.560 & 0.514 &       &       &       \\
Qwen3-32B          & 0.508 & 0.623 & 0.530 &       &       &       \\
Gemma-2-9B         & 0.522 & 0.587 & 0.518 &       &       &       \\
Gemma-2-27B        & 0.464 & 0.599 & 0.489 &       &       &       \\
MedGemma-27B-Text  & 0.517 & 0.650 & 0.546 &       &       &       \\
A.X-4.0-Light      & 0.488 & 0.585 & 0.502 &       &       &       \\
HCX-SEED-Text-1.5B & 0.499 & 0.534 & 0.483 &       &       &       \\
\bottomrule
\end{tabular}
\caption{Breakdown of factuality into precision and recall.}\label{tab:factuality_breakdown}
\end{table*}

\paragraph{Controlling for Question-Type Distribution}
To assess whether differences in question-type distributions account for the lower performance on the \textsf{Multimodal} set relative to the \textsf{Text} set, we compared performance between the two sets after controlling for question type. First, we restricted the analysis to diagnosis questions (\textsf{Text}: 1,231; \textsf{Multimodal}: 1,699). Second, we resampled 2,000 instances with replacement from the \textsf{Text} set to match the question-type distribution of the \textsf{Multimodal} set (diagnosis: 85\%, treatment: 6\%, basic veterinary knowledge: 3\%, miscellaneous: 6\%).

As shown in Table~\ref{tab:question_type_controlled}, the \textsf{Text} $>$ \textsf{Multimodal} trend remained consistent in both analyses, with all paired comparisons being statistically significant according to one-sided Wilcoxon signed-rank tests (\textit{p}$<$0.05). These analyses further support the finding in Section~\ref{subsec:zero_shot} that LVLMs underperform on the \textsf{Multimodal} set, suggesting that the observed performance gap is more likely attributable to the greater difficulty of multimodal questions than to differences in question-type distributions.

\begin{table}[t]
\centering
\small
\begin{tabular}{lcc}
\toprule
\textbf{Metric} & \textbf{Text} & \textbf{Multimodal} \\
\midrule
\multicolumn{3}{c}{\textit{Diagnosis questions only}} \\
\midrule
ROUGE       & 0.2807 & 0.2298 \\
BERTScore   & 0.7360 & 0.7196 \\
Factuality  & 0.5560 & 0.5005 \\
Helpfulness & 4.5557 & 4.3285 \\
\midrule
\multicolumn{3}{c}{\textit{Shared question type distribution}} \\
\midrule
ROUGE       & 0.2798 & 0.2303 \\
BERTScore   & 0.7353 & 0.7189 \\
Factuality  & 0.5413 & 0.4893 \\
Helpfulness & 4.5522 & 4.3157 \\
\bottomrule
\end{tabular}
\caption{Comparison of mean performance on \mydatabench after controlling for question type. The first analysis includes only diagnosis questions. The second applies the Multimodal set's question type distribution to both sets.}
\label{tab:question_type_controlled}
\end{table}

\paragraph{Factuality Breakdown into Precision and Recall}
Factuality is an F1 score that combines precision and recall to quantify the alignment between the model response and the reference answer. To better understand the relative contributions of these two components to factuality across models, we present the zero-shot breakdown in Table~\ref{tab:factuality_breakdown}. Precision is lower than recall in 26 of the 29 comparisons, with a generally larger gap on the \textsf{Text} set than on the \textsf{Multimodal} set. This trend suggests that model responses often contain information unsupported by the reference answer, even when they capture much of the information present in the reference.

\paragraph{Effects of Enhanced Reasoning}
We examined the effects of enhanced reasoning on veterinary QA performance. Table~\ref{tab:reasoning_models} shows the zero-shot performance of Qwen3 and Gemini 2.5 models with enhanced reasoning enabled. Specifically, we set \texttt{enable\_thinking=True} for Qwen3 and \texttt{thinking\_budget=256} for Gemini 2.5. The results indicate that enhanced reasoning consistently improved the helpfulness scores, whereas ROUGE, BERTScore, and factuality showed varying trends across models and modalities.

\begin{table*}[t]
\centering
\small
\setlength{\tabcolsep}{3.5pt}
\begin{tabular}{lcccc}
\toprule
Model & ROUGE & BERTScore & Factuality & Helpfulness \\
\midrule
\multicolumn{5}{c}{\textsf{Text}} \\
\midrule
Qwen3-8B & 0.294 & 0.746 & 0.514 & 3.754 \\
\quad \textit{w/ Reasoning} & 0.256 & 0.727 & 0.502 & 4.527 \\
Qwen3-32B & 0.282 & 0.742 & 0.530 & 4.433 \\
\quad \textit{w/ Reasoning} & 0.250 & 0.727 & 0.504 & 4.779 \\
\midrule
\multicolumn{5}{c}{\textsf{Multimodal}} \\
\midrule
Qwen3-VL-8B & 0.204 & 0.721 & 0.461 & 3.076 \\
\quad \textit{w/ Reasoning} & 0.197 & 0.716 & 0.484 & 3.794 \\
Gemini 2.5 Flash & 0.224 & 0.709 & 0.495 & 4.679 \\
\quad \textit{w/ Reasoning} & 0.225 & 0.714 & 0.526 & 4.767 \\
\bottomrule
\end{tabular}
\caption{Effects of enhanced reasoning on veterinary QA performance.}
\label{tab:reasoning_models}
\end{table*}

\paragraph{Benchmarking in More Languages}
We additionally conducted benchmarking experiments using the translated versions of \mydatabench in five languages: English, German, Chinese, Indonesian, and Arabic. We first validated translation quality following the protocol of \citet{doddapaneni-etal-2025-cross}. For 100 randomly sampled instances in each target language, we used GPT-4o to assign a binary validity label indicating whether each translation preserved the intended meaning without major errors. The resulting validity rates were 98\% for English and Indonesian, 97\% for German and Chinese, and 96\% for Arabic. 

We report ROUGE and BERTScore for all five languages. BERTScore was computed using \texttt{bert-base-multilingual-cased}, and ROUGE was computed using character-level tokenization for Chinese and whitespace tokenization for the other languages. We used the LLM-as-a-judge metrics only for English for two reasons. First, the factuality metric was originally proposed and tested in English~\cite{akhtar2026ev2r}. Second, LLM-as-a-judge evaluation has been studied more extensively in English than in other languages. Because we did not separately validate the helpfulness metric on the translated English version of \mydatabench, the English helpfulness results should be interpreted with caution. Future studies could validate LLM-as-a-judge metrics for other languages.

Tables~\ref{tab:zero_shot_english} and~\ref{tab:zero_shot_multilingual} present the zero-shot evaluation results across five languages. The main findings observed in Table~\ref{tab:zero_shot_results} generally persisted: closed models achieved the highest performance, and LVLMs underperformed on the \textsf{Multimodal} set in most cases. Although score differences were observed across languages, we refrain from drawing conclusions from these differences because the evaluation metrics were not implemented identically across languages.

\begin{table*}[t]
\centering
\small
\setlength{\tabcolsep}{3.5pt}
\begin{tabular}{lcccccccc}
\toprule
\multirow{2}{*}{Model} & \multicolumn{4}{c}{\textsf{Text}} & \multicolumn{4}{c}{\textsf{Multimodal}} \\
\cmidrule(lr){2-5}\cmidrule(lr){6-9}
& ROUGE & BERTScore & Factuality & Helpfulness & ROUGE & BERTScore & Factuality & Helpfulness \\
\midrule
\multicolumn{9}{c}{\textit{Closed large vision-language models}} \\
\midrule
GPT-4.1 mini & \underline{0.226} & \underline{0.725} & 0.509 & 4.886 & \underline{0.211} & \underline{0.708} & 0.473 & 4.766 \\
GPT-4.1 & 0.223 & 0.722 & \textbf{0.526} & \textbf{4.928} & 0.206 & 0.706 & \underline{0.482} & \underline{4.795} \\
Gemini 2.5 Flash & 0.211 & 0.722 & 0.507 & 4.636 & 0.191 & 0.700 & 0.448 & 4.226 \\
Gemini 2.5 Pro & 0.213 & 0.718 & \underline{0.518} & 4.912 & 0.207 & \underline{0.708} & \textbf{0.496} & 4.678 \\
\midrule
\multicolumn{9}{c}{\textit{Open-weight large vision-language models}} \\
\midrule
Qwen3-VL-8B & 0.198 & 0.716 & 0.475 & 4.378 & 0.177 & 0.700 & 0.422 & 3.837 \\
Qwen3-VL-32B & 0.210 & 0.717 & 0.483 & 4.888 & 0.195 & 0.701 & 0.440 & \textbf{4.827} \\
Gemma-3-12B & 0.190 & 0.711 & 0.496 & 4.769 & 0.189 & 0.704 & 0.456 & 4.462 \\
Gemma-3-27B & 0.184 & 0.706 & 0.485 & \underline{4.914} & 0.184 & 0.702 & 0.452 & 4.689 \\
MedGemma-27B & 0.198 & 0.715 & 0.487 & 4.757 & 0.193 & 0.705 & 0.414 & 4.370 \\
A.X-4.0-VL-Light & 0.216 & 0.724 & 0.448 & 3.988 & 0.193 & 0.707 & 0.411 & 3.927 \\
HCX-SEED-Vision-3B & 0.211 & 0.719 & 0.428 & 3.807 & \textbf{0.215} & \textbf{0.709} & 0.394 & 3.832 \\
\midrule
\multicolumn{9}{c}{\textit{Open-weight large language models}} \\
\midrule
Qwen3-8B & \textbf{0.238} & \textbf{0.730} & 0.489 & 4.157 &  &  &  & \\
Qwen3-32B & 0.217 & 0.722 & 0.492 & 4.702 &  &  &  & \\
Gemma-2-9B & 0.213 & 0.719 & 0.493 & 4.192 &  &  &  & \\
Gemma-2-27B & 0.207 & 0.717 & 0.502 & 4.304 &  &  &  & \\
MedGemma-27B-Text & 0.198 & 0.717 & 0.515 & 4.843 &  &  &  & \\
A.X-4.0-Light & 0.213 & 0.721 & 0.461 & 4.264 &  &  &  & \\
HCX-SEED-Text-1.5B & 0.217 & 0.724 & 0.422 & 3.542 &  &  &  & \\
\bottomrule
\end{tabular}
\caption{Zero-shot performance on the English version of \mydatabench. \textbf{Bold} and \underline{underlined} indicate the best and second-best results, respectively.}
\label{tab:zero_shot_english}
\end{table*}

\begin{table*}[t]
\centering
\small
\setlength{\tabcolsep}{2pt}
\renewcommand{\arraystretch}{0.92}

\begin{subtable}[t]{0.49\textwidth}
\centering
\resizebox{\linewidth}{!}{%
\begin{tabular}{@{}lcccc@{}}
\toprule
\multirow{2}{*}{Model}
& \multicolumn{2}{c}{\textsf{Text}}
& \multicolumn{2}{c}{\textsf{Multimodal}} \\
\cmidrule(lr){2-3}\cmidrule(lr){4-5}
& ROUGE & BERTScore & ROUGE & BERTScore \\
\midrule
\multicolumn{5}{c}{\textit{Closed large vision-language models}} \\
\midrule
GPT-4.1 mini       & \underline{0.235} & \textbf{0.731} & \underline{0.195} & \textbf{0.715} \\
GPT-4.1            & \textbf{0.237} & \underline{0.730} & \textbf{0.197} & \textbf{0.715} \\
Gemini 2.5 Flash   & \textbf{0.237} & 0.723 & 0.188 & 0.702 \\
Gemini 2.5 Pro     & 0.214 & 0.720 & 0.187 & 0.710 \\
\midrule
\multicolumn{5}{c}{\textit{Open-weight large vision-language models}} \\
\midrule
Qwen3-VL-8B          & 0.198 & 0.719 & 0.179 & 0.711 \\
Qwen3-VL-32B         & 0.212 & 0.725 & 0.182 & 0.711 \\
Gemma-3-12B          & 0.218 & 0.722 & 0.192 & 0.710 \\
Gemma-3-27B          & 0.220 & 0.724 & \underline{0.195} & \underline{0.714} \\
MedGemma-27B         & 0.223 & 0.721 & 0.190 & 0.706 \\
A.X-4.0-VL-Light     & 0.211 & 0.719 & 0.187 & 0.709 \\     
HCX-SEED-Vision-3B   & 0.205 & 0.707 & 0.182 & 0.689 \\
\midrule
\multicolumn{5}{c}{\textit{Open-weight large language models}} \\
\midrule
Qwen3-8B             & 0.231 & \textbf{0.731} &       &       \\
Qwen3-32B            & 0.224 & 0.728 &       &       \\
Gemma-2-9B           & 0.227 & 0.723 &       &       \\
Gemma-2-27B          & 0.218 & 0.718 &       &       \\
MedGemma-27B-Text    & 0.225 & 0.725 &       &       \\
A.X-4.0-Light        & 0.217 & 0.722 &       &       \\
HCX-SEED-Text-1.5B   & 0.199 & 0.693 &       &       \\
\bottomrule
\end{tabular}%
}
\caption{Chinese}
\label{tab:zero_shot_chinese}
\end{subtable}
\hfill
\begin{subtable}[t]{0.49\textwidth}
\centering
\resizebox{\linewidth}{!}{%
\begin{tabular}{@{}lcccc@{}}
\toprule
\multirow{2}{*}{Model}
& \multicolumn{2}{c}{\textsf{Text}}
& \multicolumn{2}{c}{\textsf{Multimodal}} \\
\cmidrule(lr){2-3}\cmidrule(lr){4-5}
& ROUGE & BERTScore & ROUGE & BERTScore \\
\midrule
\multicolumn{5}{c}{\textit{Closed large vision-language models}} \\
\midrule
GPT-4.1 mini       & \underline{0.209} & \underline{0.713} & 0.196 & \textbf{0.697} \\
GPT-4.1            & 0.200 & 0.708 & 0.187 & 0.693 \\
Gemini 2.5 Flash   & \underline{0.209} & 0.711 & 0.195 & 0.689 \\
Gemini 2.5 Pro     & 0.198 & 0.704 & 0.196 & 0.694 \\
\midrule
\multicolumn{5}{c}{\textit{Open-weight large vision-language models}} \\
\midrule
Qwen3-VL-8B          & 0.191 & 0.706 & 0.191 & \underline{0.695} \\
Qwen3-VL-32B         & 0.195 & 0.706 & 0.180 & 0.689 \\
Gemma-3-12B          & 0.196 & 0.703 & \underline{0.197} & 0.691 \\
Gemma-3-27B          & 0.198 & 0.706 & \textbf{0.200} & 0.692 \\
MedGemma-27B         & 0.197 & 0.705 & 0.194 & 0.691 \\
A.X-4.0-VL-Light     & 0.190 & 0.695 & 0.188 & 0.690 \\   
HCX-SEED-Vision-3B   & 0.186 & 0.687 & 0.171 & 0.670 \\
\midrule
\multicolumn{5}{c}{\textit{Open-weight large language models}} \\
\midrule
Qwen3-8B             & \textbf{0.213} & \textbf{0.715} &       &       \\
Qwen3-32B            & 0.196 & 0.707 &       &       \\
Gemma-2-9B           & 0.206 & 0.704 &       &       \\
Gemma-2-27B          & 0.202 & 0.700 &       &       \\
MedGemma-27B-Text    & 0.204 & 0.709 &       &       \\
A.X-4.0-Light        & 0.203 & 0.706 &       &       \\
HCX-SEED-Text-1.5B   & 0.174 & 0.651 &       &       \\
\bottomrule
\end{tabular}%
}
\caption{German}
\label{tab:zero_shot_german}
\end{subtable}

\par\medskip

\begin{subtable}[t]{0.49\textwidth}
\centering
\resizebox{\linewidth}{!}{%
\begin{tabular}{@{}lcccc@{}}
\toprule
\multirow{2}{*}{Model}
& \multicolumn{2}{c}{\textsf{Text}}
& \multicolumn{2}{c}{\textsf{Multimodal}} \\
\cmidrule(lr){2-3}\cmidrule(lr){4-5}
& ROUGE & BERTScore & ROUGE & BERTScore \\
\midrule
\multicolumn{5}{c}{\textit{Closed large vision-language models}} \\
\midrule
GPT-4.1 mini       & \textbf{0.229} & \underline{0.730} & \textbf{0.194} & \textbf{0.710} \\
GPT-4.1            & 0.221 & 0.727 & 0.186 & \textbf{0.710} \\
Gemini 2.5 Flash   & 0.218 & 0.724 & 0.186 & 0.701 \\
Gemini 2.5 Pro     & 0.211 & 0.720 & 0.189 & \underline{0.708} \\
\midrule
\multicolumn{5}{c}{\textit{Open-weight large vision-language models}} \\
\midrule
Qwen3-VL-8B          & 0.200 & 0.726 & 0.170 & \underline{0.708} \\
Qwen3-VL-32B         & 0.210 & 0.725 & 0.184 & 0.707 \\
Gemma-3-12B          & 0.206 & 0.719 & 0.187 & 0.707 \\
Gemma-3-27B          & 0.206 & 0.720 & \underline{0.190} & 0.707 \\
MedGemma-27B         & 0.206 & 0.720 & 0.184 & 0.702 \\
A.X-4.0-VL-Light     & 0.181 & 0.712 & 0.171 & 0.705 \\  
HCX-SEED-Vision-3B   & 0.167 & 0.691 & 0.138 & 0.666 \\
\midrule
\multicolumn{5}{c}{\textit{Open-weight large language models}} \\
\midrule
Qwen3-8B             & \underline{0.222} & \textbf{0.731} &       &       \\
Qwen3-32B            & 0.211 & 0.725 &       &       \\
Gemma-2-9B           & 0.216 & 0.721 &       &       \\
Gemma-2-27B          & 0.210 & 0.717 &       &       \\
MedGemma-27B-Text    & 0.213 & 0.723 &       &       \\
A.X-4.0-Light        & 0.193 & 0.716 &       &       \\
HCX-SEED-Text-1.5B   & 0.124 & 0.650 &       &       \\
\bottomrule
\end{tabular}%
}
\caption{Indonesian}
\label{tab:zero_shot_indonesian}
\end{subtable}
\hfill
\begin{subtable}[t]{0.49\textwidth}
\centering
\resizebox{\linewidth}{!}{%
\begin{tabular}{@{}lcccc@{}}
\toprule
\multirow{2}{*}{Model}
& \multicolumn{2}{c}{\textsf{Text}}
& \multicolumn{2}{c}{\textsf{Multimodal}} \\
\cmidrule(lr){2-3}\cmidrule(lr){4-5}
& ROUGE & BERTScore & ROUGE & BERTScore \\
\midrule
\multicolumn{5}{c}{\textit{Closed large vision-language models}} \\
\midrule
GPT-4.1 mini       & \textbf{0.175} & \textbf{0.728} & 0.155 & \textbf{0.715} \\
GPT-4.1            & \underline{0.171} & \underline{0.726} & 0.154 & 0.713 \\
Gemini 2.5 Flash   & 0.168 & 0.725 & 0.155 & 0.707 \\
Gemini 2.5 Pro     & 0.152 & 0.720 & 0.148 & 0.712 \\
\midrule
\multicolumn{5}{c}{\textit{Open-weight large vision-language models}} \\
\midrule
Qwen3-VL-8B          & 0.150 & 0.717 & 0.132 & 0.707 \\
Qwen3-VL-32B         & 0.158 & 0.722 & 0.135 & 0.708 \\
Gemma-3-12B          & 0.160 & 0.722 & \underline{0.159} & \underline{0.714} \\
Gemma-3-27B          & 0.162 & 0.723 & \textbf{0.161} & \underline{0.714} \\
MedGemma-27B         & 0.160 & 0.721 & 0.151 & 0.709 \\
A.X-4.0-VL-Light     & 0.149 & 0.702 & 0.134 & 0.689 \\  
HCX-SEED-Vision-3B   & 0.141 & 0.661 & 0.100 & 0.545 \\
\midrule
\multicolumn{5}{c}{\textit{Open-weight large language models}} \\
\midrule
Qwen3-8B             & \textbf{0.175} & \textbf{0.728} &       &       \\
Qwen3-32B            & 0.170 & 0.725 &       &       \\
Gemma-2-9B           & 0.170 & 0.722 &       &       \\
Gemma-2-27B          & 0.166 & 0.718 &       &       \\
MedGemma-27B-Text    & 0.159 & 0.722 &       &       \\
A.X-4.0-Light        & 0.147 & 0.697 &       &       \\
HCX-SEED-Text-1.5B   & 0.127 & 0.678 &       &       \\
\bottomrule
\end{tabular}%
}
\caption{Arabic}
\label{tab:zero_shot_arabic}
\end{subtable}

\caption{Zero-shot performance on the Chinese, German, Indonesian, and Arabic versions of \mydatabench. \textbf{Bold} and \underline{underlined} indicate the best and second-best results, respectively.}
\label{tab:zero_shot_multilingual}
\end{table*}

\begin{table*}[t]
\centering
\small
\resizebox{\textwidth}{!}{
\begin{tabular}{lcccccccc}
\toprule
\multirow{2}{*}{Model} & \multicolumn{4}{c}{\textsf{Text}} & \multicolumn{4}{c}{\textsf{Multimodal}} \\
\cmidrule(lr){2-5} \cmidrule(lr){6-9}
& ROUGE & BERTScore & Factuality & Helpfulness & ROUGE & BERTScore & Factuality & Helpfulness \\
\midrule

\multicolumn{9}{c}{\textit{Closed large vision-language models}} \\
\midrule
GPT-4.1 mini & \textbf{0.299}\down{0.004} & \underline{0.745}\down{0.001} & \underline{0.569}\up{0.018} & 4.827\up{0.099} & 0.235\down{0.002} & \underline{0.724}\down{0.002} & \underline{0.534}\up{0.003} & 4.780\up{0.183} \\
GPT-4.1 & 0.287\down{0.003} & 0.741\down{0.002} & \textbf{0.571}\up{0.002} & \textbf{4.931}\up{0.051} & 0.228\down{0.006} & 0.722\down{0.004} & 0.531\same & \underline{4.809}\up{0.094} \\
Gemini 2.5 Flash & 0.276\down{0.003} & 0.729\down{0.002} & 0.536\down{0.005} & 4.777\down{0.137} & 0.221\down{0.003} & 0.702\down{0.007} & 0.497\up{0.002} & 4.727\up{0.048} \\
Gemini 2.5 Pro & 0.292\up{0.012} & 0.741\up{0.003} & 0.556\up{0.006} & \underline{4.921}\down{0.002} & \textbf{0.248}\up{0.002} & \textbf{0.727}\up{0.001} & \textbf{0.539}\up{0.015} & \textbf{4.866}\up{0.052} \\

\midrule
\multicolumn{9}{c}{\textit{Open-weight large vision-language models}} \\
\midrule
Qwen3-VL-8B & 0.279\up{0.008} & 0.739\up{0.002} & 0.508\up{0.009} & 3.986\up{0.025} & 0.227\up{0.023} & 0.722\up{0.001} & 0.479\up{0.018} & 3.460\up{0.384} \\
Qwen3-VL-32B & 0.252\down{0.006} & 0.726\down{0.004} & 0.522\up{0.009} & 4.825\down{0.042} & 0.209\down{0.006} & 0.704\down{0.008} & 0.485\down{0.004} & 4.721\down{0.073} \\
Gemma-3-12B & 0.277\down{0.010} & 0.732\down{0.008} & 0.503\down{0.007} & 4.434\down{0.127} & 0.232\down{0.011} & 0.717\down{0.006} & 0.457\down{0.033} & 4.032\down{0.223} \\
Gemma-3-27B & 0.276\down{0.004} & 0.734\down{0.003} & 0.523\up{0.004} & 4.815\down{0.055} & \underline{0.237}\down{0.007} & 0.721\down{0.003} & 0.506\up{0.010} & 4.641\up{0.015} \\
MedGemma-27B & 0.256\down{0.003} & 0.716\up{0.001} & 0.504\up{0.001} & 4.791\down{0.081} & 0.216\down{0.004} & 0.700\down{0.002} & 0.454\up{0.012} & 4.591\down{0.043} \\
A.X-4.0-VL-Light & 0.289\up{0.012} & 0.734\up{0.007} & 0.472\down{0.010} & 3.510\down{0.221} & \underline{0.237}\down{0.001} & 0.717\down{0.005} & 0.463\up{0.002} & 3.882\up{0.259} \\
HCX-SEED-Vision-3B & 0.281\down{0.003} & 0.732\down{0.003} & 0.495\up{0.002} & 3.896\up{0.120} & 0.230\same & 0.711\down{0.007} & 0.467\up{0.016} & 3.942\up{0.271} \\

\midrule
\multicolumn{9}{c}{\textit{Open-weight large language models}} \\
\midrule
Qwen3-8B & \underline{0.294}\same & \textbf{0.746}\same & 0.508\down{0.006} & 3.801\up{0.047} & & & & \\
Qwen3-32B & 0.281\down{0.001} & 0.741\down{0.001} & 0.534\up{0.004} & 4.524\up{0.091} & & & & \\
Gemma-2-9B & \textbf{0.299}\up{0.011} & 0.744\up{0.011} & 0.502\down{0.016} & 3.615\down{0.561} & & & &  \\
Gemma-2-27B & 0.291\up{0.016} & 0.730\up{0.014} & 0.510\up{0.021} & 4.296\down{0.258} & & & &  \\
MedGemma-27B-Text & 0.276\down{0.012} & 0.732\down{0.007} & 0.528\down{0.018} & 4.565\down{0.229} & & & &  \\
A.X-4.0-Light & 0.274\down{0.015} & 0.723\down{0.007} & 0.503\up{0.001} & 4.561\up{0.349} & & & &  \\
HCX-SEED-Text-1.5B & 0.281\down{0.014} & 0.733\down{0.010} & 0.487\up{0.004} & 3.707\up{0.228} & & & & \\

\bottomrule
\end{tabular}
}
\caption{Performance of RAG on \mydatabench. Parenthetical values report $\Delta=\text{RAG}-\text{zero-shot}$, with blue text indicating positive values and red text indicating a negative value. \textbf{Bold} and \underline{underlined} indicate the best and second-best results, respectively.}
\label{tab:rag_results}
\end{table*}

\begin{table*}[t]
\centering
\small
\resizebox{\textwidth}{!}{
\begin{tabular}{lcccccccc}
\toprule
\multirow{2}{*}{Model} & \multicolumn{4}{c}{\textsf{Text}} & \multicolumn{4}{c}{\textsf{Multimodal}} \\
\cmidrule(lr){2-5} \cmidrule(lr){6-9}
& ROUGE & BERTScore & Factuality & Helpfulness & ROUGE & BERTScore & Factuality & Helpfulness \\
\midrule
\multicolumn{9}{c}{\textit{Open-weight large vision-language models}} \\
\midrule
Qwen3-VL-8B
& 0.341\up{0.070} & 0.771\up{0.034} & 0.477\down{0.022} & 2.649\down{1.312}
& 0.239\up{0.035} & 0.698\down{0.023} & 0.317\down{0.144} & 1.978\down{1.098} \\
Qwen3-VL-32B
& 0.354\up{0.096} & 0.776\up{0.046} & 0.513\same & 2.880\down{1.987}
& 0.266\up{0.051} & 0.724\up{0.012} & 0.379\down{0.110} & 2.343\down{2.451} \\
Gemma-3-12B
& 0.360\up{0.073} & 0.775\up{0.035} & 0.498\down{0.012} & 2.788\down{1.773}
& 0.299\up{0.056} & 0.745\up{0.022} & 0.409\down{0.081} & 2.463\down{1.792} \\
Gemma-3-27B
& \textbf{0.366}\up{0.086} & 0.777\up{0.040} & 0.507\down{0.012} & \underline{2.897}\down{1.973}
& \underline{0.321}\up{0.077} & \underline{0.757}\up{0.033} & \underline{0.433}\down{0.063} & 2.499\down{2.127} \\
MedGemma-27B
& 0.361\up{0.102} & \underline{0.779}\up{0.064} & \underline{0.518}\up{0.015} & 2.878\down{1.994}
& \textbf{0.324}\up{0.104} & \textbf{0.763}\up{0.061} & \textbf{0.476}\up{0.034} & \textbf{2.694}\down{1.940} \\
A.X-4.0-VL-Light
& 0.327\up{0.050} & 0.759\up{0.032} & 0.431\down{0.051} & 2.392\down{1.339}
& 0.261\up{0.023} & 0.740\up{0.018} & 0.384\down{0.077} & \underline{2.502}\down{1.121} \\
HCX-SEED-Vision-3B
& 0.341\up{0.057} & 0.771\up{0.036} & 0.469\down{0.024} & 2.571\down{1.205}
& 0.280\up{0.050} & 0.747\up{0.029} & 0.399\down{0.052} & 2.402\down{1.269} \\

\midrule
\multicolumn{9}{c}{\textit{Open-weight large language models}} \\
\midrule
Qwen3-8B
& 0.337\up{0.043} & 0.769\up{0.023} & 0.471\down{0.043} & 2.629\down{1.125}
& & & & \\
Qwen3-32B
& 0.348\up{0.066} & 0.775\up{0.033} & 0.506\down{0.024} & 2.849\down{1.584}
& & & & \\
Gemma-2-9B
& 0.348\up{0.060} & 0.776\up{0.043} & 0.506\down{0.012} & 2.736\down{1.440}
& & & & \\
Gemma-2-27B
& 0.350\up{0.075} & 0.776\up{0.060} & 0.517\up{0.028} & 2.787\down{1.767}
& & & & \\
MedGemma-27B-Text
& \underline{0.362}\up{0.074} & \textbf{0.780}\up{0.041} & \textbf{0.524}\down{0.022} & \textbf{2.943}\down{1.851}
& & & & \\
A.X-4.0-Light
& 0.346\up{0.057} & 0.773\up{0.043} & 0.486\down{0.016} & 2.794\down{1.418}
& & & & \\
HCX-SEED-Text-1.5B
& 0.336\up{0.041} & 0.769\up{0.026} & 0.451\down{0.032} & 2.466\down{1.013}
& & & & \\

\bottomrule
\end{tabular}
}
\caption{SFT performance measured on \mydatabench. Parenthetical values report $\Delta=\text{SFT}-\text{zero-shot}$, with blue text indicating positive values and red text indicating negative values. \textbf{Bold} and \underline{underlined} indicate the best and second-best results, respectively.}
\label{tab:sft_results}
\end{table*}

\section{Prompts}
\label{appendix:prompts}
This section presents the prompts used in our experiments. All prompts were originally written in Korean, except for the judge prompts used to evaluate the English version of \mydatabench, which were written in English. For accessibility, we provide English translations of the prompts used in our experiments in this manuscript, while the original Korean prompts are available in our GitHub repository. Figure~\ref{fig:llm_preprocessing_prompts_en} shows the prompt used for LLM-based preprocessing (Section~\ref{sec:data_collection_preprocessing}). The LLM-as-a-judge prompts for factuality and helpfulness evaluation are presented in Figures~\ref{fig:factuality_judge_prompt_en} and~\ref{fig:helpfulness_judge_prompt_en}. Figures~\ref{fig:zero_shot_prompt_en} and \ref{fig:rag_prompt_en} show the prompts used for the zero-shot and RAG settings, respectively. We used the same zero-shot prompt for SFT. 

\begin{figure*}[t]
    \centering
    \begin{tcolorbox}[
        enhanced,
        title=\textbf{Prompt: Preprocessing (Filtering)},
        arc=2mm,
        boxrule=0.5pt,
        fontupper=\small
    ]
    Your role is to filter out irrelevant data to construct a veterinary medical question answering dataset for pets (dogs and cats). Please refer to the filtering criteria below and return strictly ``true'' or ``false''. Do not include any other text in your response. \\
    
    \textbf{\#\#\# Filtering Criteria} \\
    Return ``false'' if any of the following conditions apply: \\
    - Questions unrelated to veterinary medical consultations for pets (dogs and cats). \\
    - Questions regarding admission to veterinary schools, the veterinary profession, or licensing. \\
    - Questions containing personally identifiable information (PII) or content that violates service operation policies. \\
    - Answers based on groundless speculation. \\
    - Answers with little to no informational value, such as simply stating ``Please visit a veterinary clinic.'' \\
    
    If none of the above conditions apply, return ``true''. \\
    
    \textbf{Question:} \texttt{\{question\}} \\
    \textbf{Answer:} \texttt{\{answer\}}
    \end{tcolorbox}
    
    \vspace{0.5cm}
    
    \begin{tcolorbox}[
        enhanced,
        title=\textbf{Prompt: Preprocessing (Cleaning)},
        arc=2mm,
        boxrule=0.5pt,
        fontupper=\small
    ]
    Your role is to remove unnecessary phrases and refine expressions to construct a high-quality veterinary medical question answering dataset for pets (dogs and cats). Please refer to the cleaning guidelines below to preprocess the given data into a ``preprocessed question'' and ``preprocessed answer''. You must output the result strictly in the JSON format specified below. \\
    
    \textbf{\#\#\# Cleaning Guidelines} \\
    Maintain the core meaning of the original question and answer while removing the following elements. Ensure that critical details such as pet demographics (species, age, weight) and clinical symptoms are not omitted from the question. Similarly, ensure that essential information, such as diagnostic procedures and treatment methods, is preserved in the answer. \\
    - Remove Personally Identifiable Information (PII): Delete names, phone numbers, email addresses, physical addresses, etc. \\
    - Correct Spelling and Grammar: Rectify errors considering the context. \\
    - Convert Colloquialisms: Transform conversational language into a formal, written style appropriate for official Q\&A formats. \\
    - Remove Promotional Content: Delete advertisements or promotional phrases. \\
    - Remove Unnecessary Emotional Expressions: Delete phrases that do not contribute to the factual meaning. \\
    - Remove Other Superfluous Phrases: Delete unnecessary conversational fillers and greetings. \\

    \textbf{\#\#\# Output Format} \\
    \begin{verbatim}
    {
        "preprocessed_question": "string",
        "preprocessed_answer": "string"
    }
    \end{verbatim}
    
    \textbf{Question:} \texttt{\{question\}} \\
    \textbf{Answer:} \texttt{\{answer\}}
    \end{tcolorbox}
    \caption{LLM-based preprocessing prompt.}
    \label{fig:llm_preprocessing_prompts_en}
\end{figure*}

\begin{figure*}[t]
    \centering
    \begin{tcolorbox}[
        enhanced,
        title=\textbf{Prompt: LLM-as-a-Judge (Factuality)},
        arc=2mm,
        boxrule=0.5pt,
        fontupper=\small
    ]
    You will be given a question, a reference answer, and a predicted answer.
    Please verify the factual correctness of the predicted answer by comparing it with the reference answer, following these steps: \\
    
    1. Break down the predicted answer into independent factual statements. Each fact should be a separate sentence. \\
    2. Evaluate each fact individually: determine whether the reference answer supports the fact. Do not use external knowledge or additional background knowledge. \\
    3. Next, break down the reference answer into independent factual statements. Each fact should be a separate sentence. \\
    4. Evaluate each fact individually: determine whether the predicted answer supports the fact. Do not use external knowledge or additional background knowledge. \\
    5. Finally, summarize (1.) how many predicted facts the reference answer supports and (2.) how many reference facts are supported by the predicted answer. \\
    
    Generate the output in JSON format as shown in the examples below. \\
   \texttt{\{Example 1\}} \\
   \texttt{\{Example 2\}} \\
    
    \textbf{\#\#\# Input} \\
    \textbf{Question:} \texttt{\{question\}} \\
    \textbf{Reference Answer:} \texttt{\{gold\_answer\}} \\
    \textbf{Predicted Answer:} \texttt{\{pred\_answer\}} \\
    \textbf{Output:}
    \end{tcolorbox}
    \caption{LLM-as-a-judge prompt used for factuality evaluation.}
    \label{fig:factuality_judge_prompt_en}
\end{figure*}

\begin{figure*}[t]
    \centering
    \begin{tcolorbox}[
        enhanced,
        title=\textbf{Prompt: LLM-as-a-Judge (Helpfulness)},
        arc=2mm,
        boxrule=0.5pt,
        fontupper=\small
    ]

    You are an expert at evaluating the quality of text. \\
    As an impartial evaluator, please assess how useful the AI assistant's response is to the user's question. \\
    Specifically, evaluate the following criteria: 
    1) relevance to the question; 
    2) whether it meets the user's purpose and needs; 
    3) whether it provides a sufficient and appropriate answer. \\
    You must first provide an analysis, and then strictly assign a score between 1 and 5 in the following format: ``[[Rating]]'', e.g., ``[[5]]''. \\

    \textbf{\#\#\# Examples} \\
    \texttt{\{Example 1\}} \\
    \texttt{\{Example 2\}} \\
    \texttt{\{Example 3\}} \\

    Now, please evaluate the following AI assistant's response based on the evaluation criteria and examples above: \\

    \textbf{Question} \\
    \texttt{\{question\}} \\
    \textbf{Assistant's Answer Begins} \\
    \texttt{\{pred\_answer\}} \\
    \textbf{Assistant's Answer Ends} \\
    \textbf{Analysis}
    \end{tcolorbox}
    \caption{LLM-as-a-judge prompt used for helpfulness evaluation.}
    \label{fig:helpfulness_judge_prompt_en}
\end{figure*}

\begin{figure*}[t]
    \centering
    \begin{tcolorbox}[
        enhanced,
        title=\textbf{Prompt: Question Answering (Zero-Shot)},
        arc=2mm,
        boxrule=0.5pt,
        fontupper=\small
    ]
    Your role is to answer the given question. 
    Do not generate more than five sentences in your response. \\

    \textbf{Question:} \texttt{\{question\}} \\
    \textbf{Answer:}
    \end{tcolorbox}
    \caption{Zero-shot prompt for question answering. The same prompt is used under supervised fine-tuning.}
    \label{fig:zero_shot_prompt_en}
\end{figure*}

\begin{figure*}[t]
    \centering
    \begin{tcolorbox}[
        enhanced,
        title=\textbf{Prompt: Question Answering (RAG)},
        arc=2mm,
        boxrule=0.5pt,
        fontupper=\small
    ]
    Your role is to answer the given question. Do not generate more than five sentences in your response. \\
    Retrieved documents are provided below. If the retrieved documents are relevant to the question, refer to them when answering; otherwise, ignore them and answer based on your own knowledge. \\

    \textbf{Question:} \texttt{\{question\}} \\
    \textbf{Retrieved Documents:} \texttt{\{paragraphs\}} \\
    \textbf{Answer:}
    \end{tcolorbox}
    \caption{Prompt used for retrieval-augmented generation (RAG).}
    \label{fig:rag_prompt_en}
\end{figure*}

\section{Human Validation}

\paragraph{LLM-based Preprocessing}
\label{appendix:llm-based_preprocessing}

We manually evaluated 100 randomly sampled QA pairs to validate the LLM-based preprocessing. One author compared each original and preprocessed pair using two criteria: \emph{coherence}, whether the preprocessed output remained logical, grammatical, and natural; and \emph{completeness}, whether essential clinical information, including symptoms, diseases, and pet metadata was preserved.

\paragraph{Question Type and Clinical Condition}

\begin{figure*}[t]
    \centering
    \begin{subfigure}[b]{0.48\textwidth}
        \centering
        \includegraphics[width=\linewidth]
            {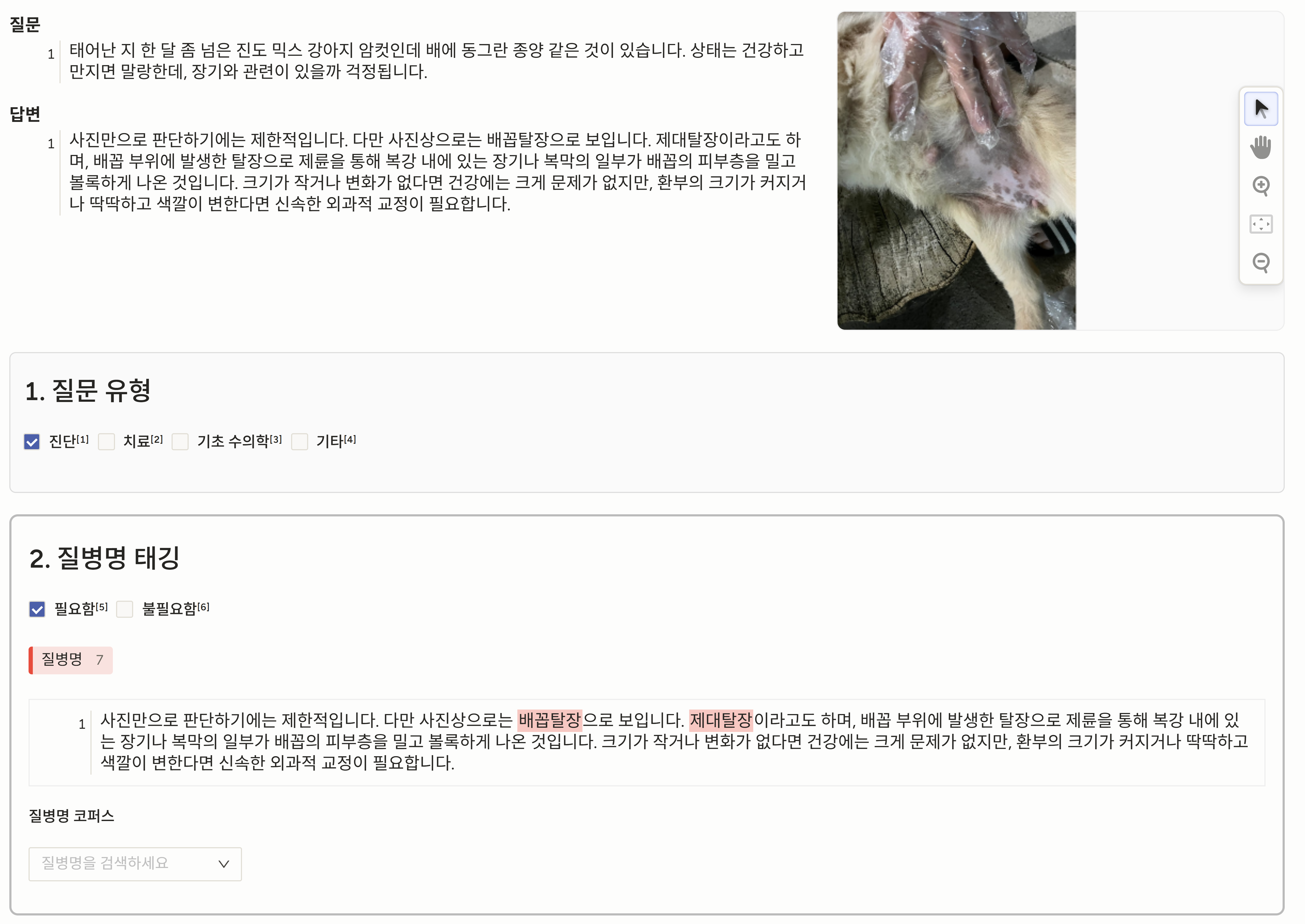}
        \caption{For question types and clinical conditions.}
        \label{fig:annotator_interface}
    \end{subfigure}%
    \hfill%
    \begin{subfigure}[b]{0.48\textwidth}
        \centering
        \includegraphics[width=\linewidth]
            {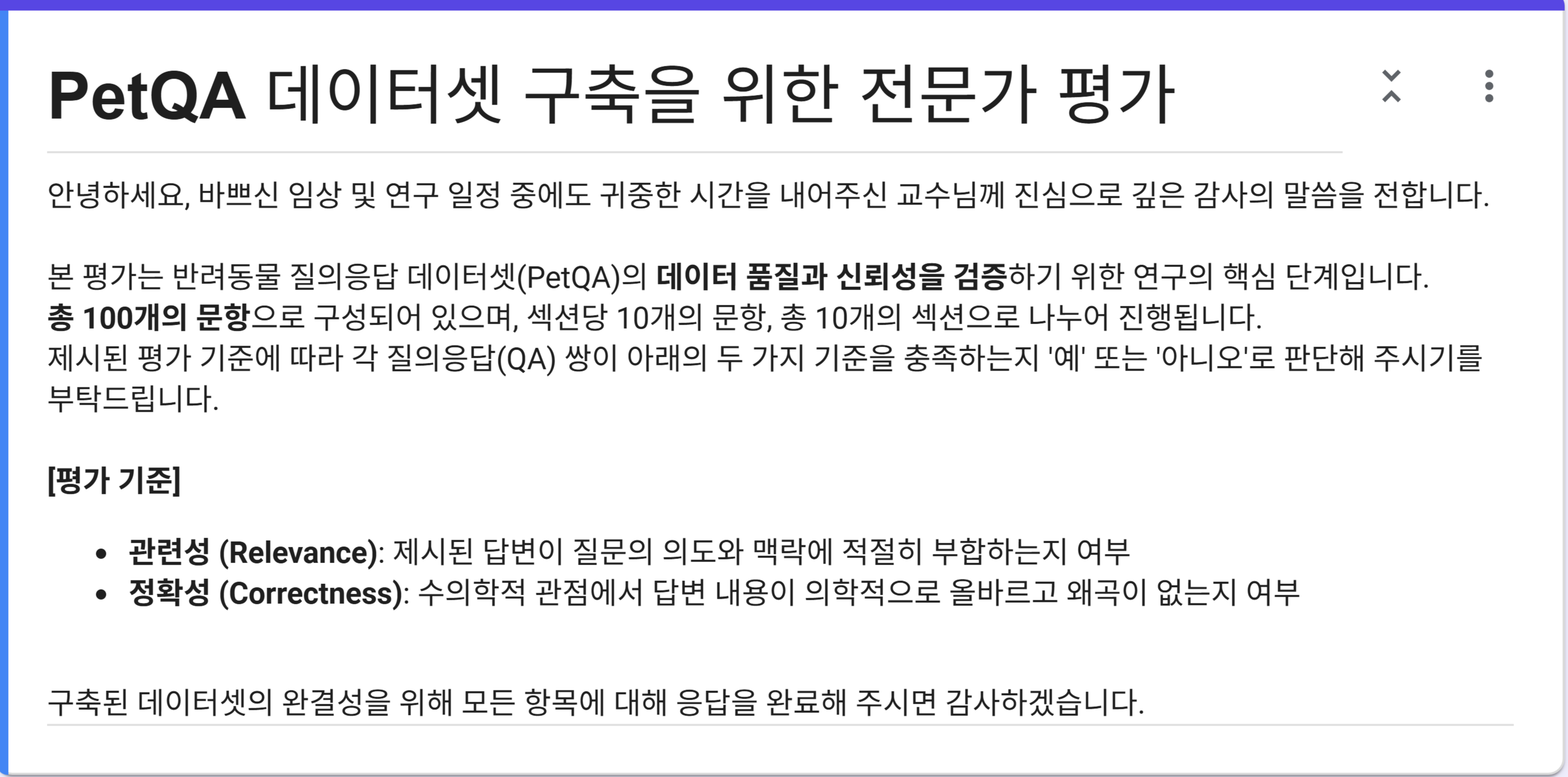}

        \vspace{0.5em}

        \includegraphics[width=\linewidth]
            {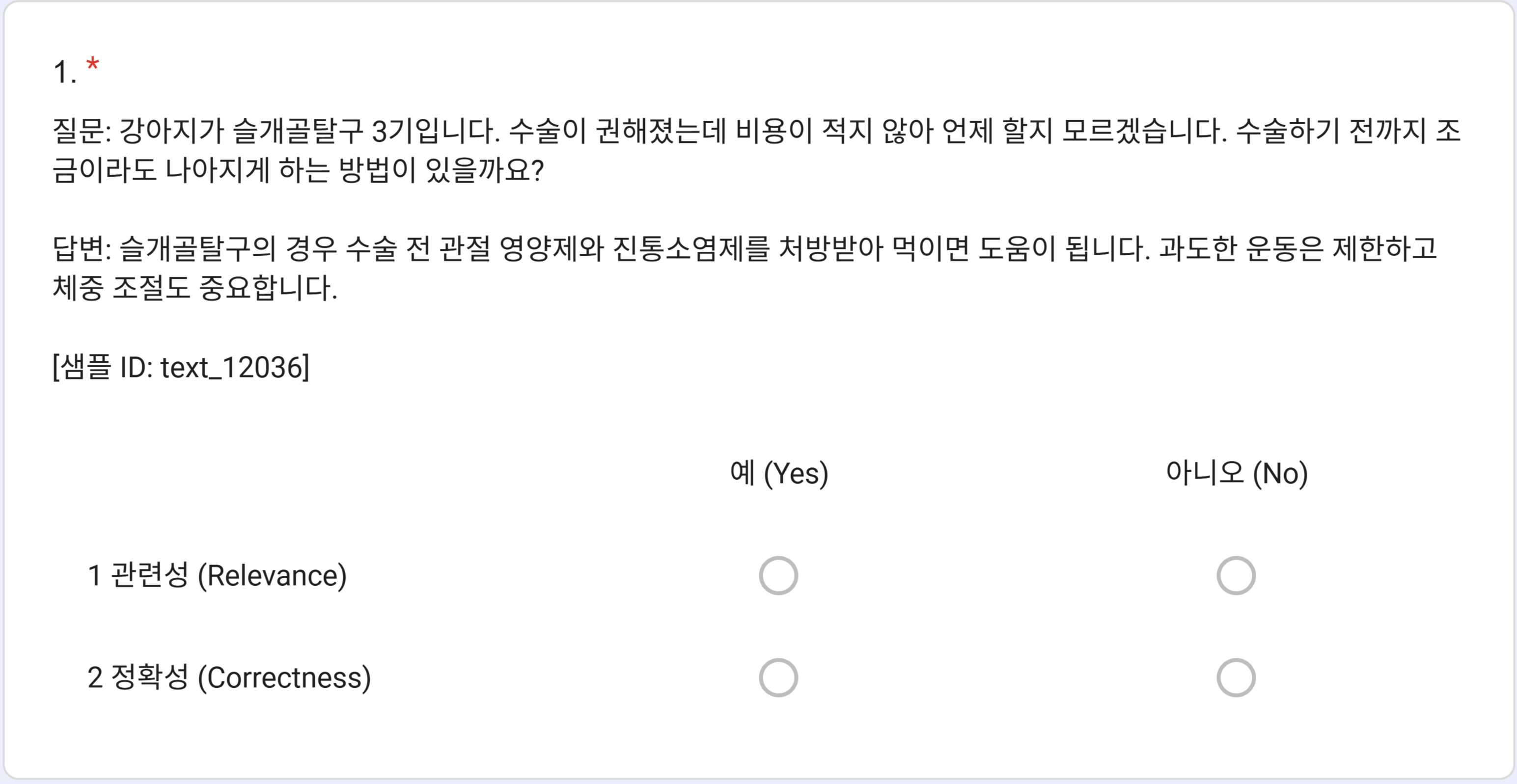}
        \caption{For expert verification of dataset quality.}
        \label{fig:expert_validation_interface}
    \end{subfigure}

    \caption{Interfaces used for dataset annotation and expert validation.}
    \label{fig:annotation_validation_interfaces}
\end{figure*}

Figure~\ref{fig:annotator_interface} shows the interface used to assign a question type and, for \emph{Diagnosis} questions, to annotate clinical conditions in the selected answer. Exact string matches were automatically pre-annotated, and annotators manually added conditions missed by the matching procedure due to surface-form variations, including differences in spacing, spelling, and suffixes.

\paragraph{Expert Verification}
\label{appendix:expert_verification}

Using the interface shown in Figure~\ref{fig:expert_validation_interface}, the expert identified factual errors and potentially misleading clinical statements as the primary causes of incorrect cases. The observed correctness rate was comparable to the 86\% expert-verified accuracy reported in a prior medical QA benchmark~\cite{zhang-etal-2025-llmeval}. We additionally assessed the sensitivity of model rankings by comparing results computed using all audited references with those computed using only the medically correct subset. The Spearman rank correlations were \(0.936\) for ROUGE and factuality, \(0.973\) for BERTScore, and \(0.975\) for helpfulness. These high correlations suggest that excluding incorrect references did not affect the overall ranking.

\end{document}